\documentclass[10pt,twocolumn,letterpaper]{article}

\usepackage[pagenumbers]{cvpr} 

\usepackage{multirow}

\definecolor{cvprblue}{rgb}{0.21,0.49,0.74}
\usepackage[pagebackref,breaklinks,colorlinks,allcolors=cvprblue]{hyperref}

\usepackage{booktabs}
\usepackage{adjustbox}

\newcommand\blfootnote[1]{%
  \begingroup
  \renewcommand\thefootnote{}\footnote{#1}%
  \addtocounter{footnote}{-1}%
  \endgroup
}

\title{StyleDiT: A Unified Framework for Diverse Child and Partner Faces Synthesis with Style Latent Diffusion Transformer}

\author{Pin-Yen Chiu$^{*}$ \hspace{0.9cm}
Dai-Jie Wu$^{*}$ \hspace{0.9cm}
Po-Hsun Chu \\
Chia-Hsuan Hsu \hspace{0.9cm}
Hsiang-Chen Chiu \hspace{0.9cm}
Chih-Yu Wang \hspace{0.9cm}
Jun-Cheng Chen
\vspace{0.2cm} \\
Research Center for Information Technology Innovation, Academia Sinica\\
{\tt\small nickchiu@citi.sinica.edu.tw, bbb50111@gmail.com, benny0808@citi.sinica.edu.tw}\\
{\tt\small \{winnie920924, charly729.chiu\}@gmail.com, \{cywang, pullpull\}@citi.sinica.edu.tw}
}

\begin{document}

\twocolumn[{%
\vspace{-10mm}
\maketitle
\vspace{-8mm}
\begin{center}
    \centering
    \captionsetup{type=figure}
    \includegraphics[width=\linewidth]{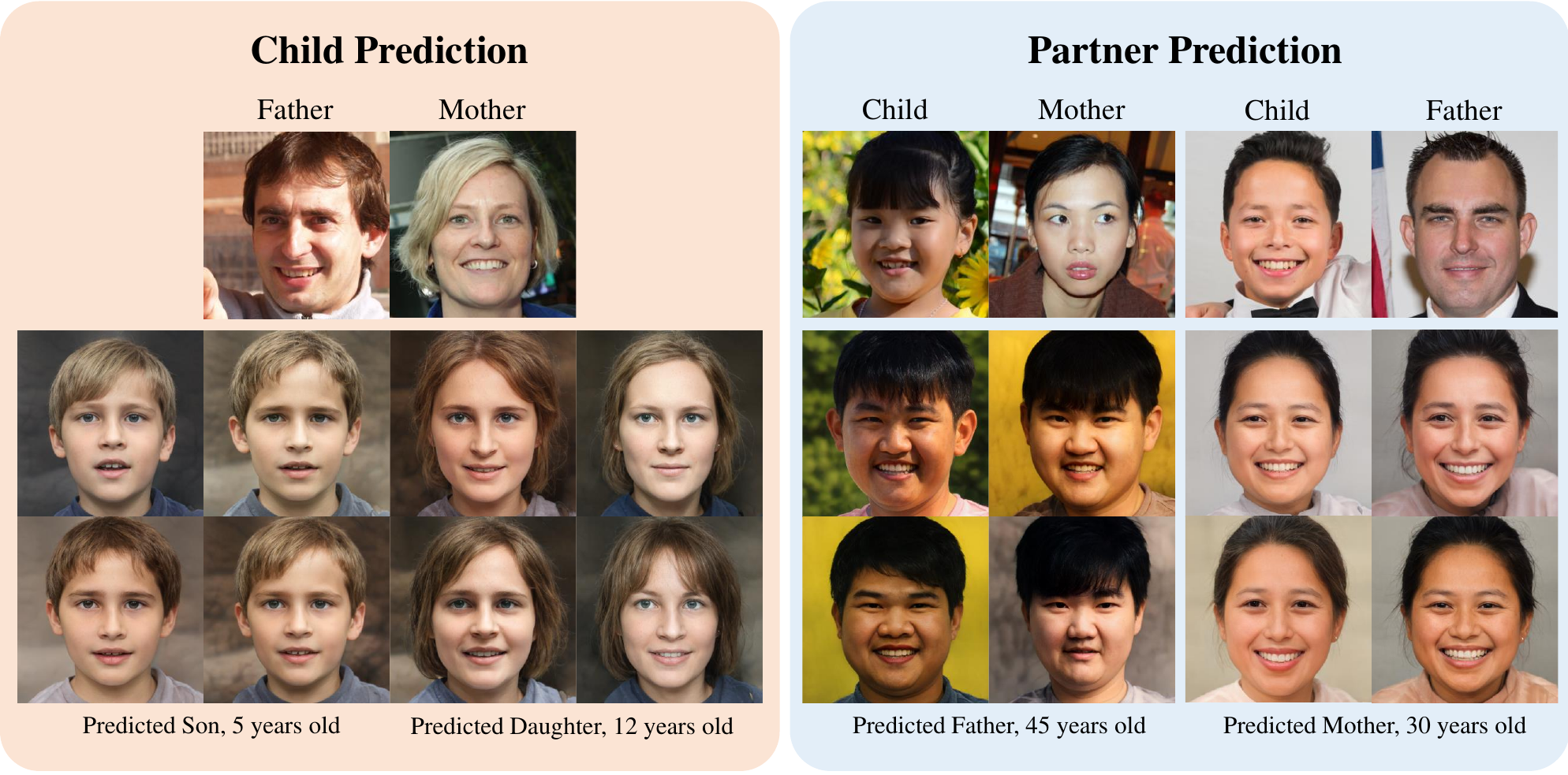}
    \vspace{-4mm}
    \captionof{figure}{StyleDiT showcases dual capabilities: it synthesizes child faces from parents' images and generates partner faces from a child's and one parent's face. For child prediction, it produces diverse child faces from parents, accommodating different ages and genders. Similarly, in partner prediction, it effectively creates varied parental images using a child's and one parent's face.}\label{fig:teaser}
\end{center}%
}]
\begin{abstract}
\blfootnote{*indicates equal contribution.}
Kinship face synthesis is a challenging problem due to the scarcity and low quality of the available kinship data.
Existing methods often struggle to generate descendants with both high diversity and fidelity while precisely controlling facial attributes such as age and gender.
To address these issues, we propose the Style Latent Diffusion Transformer (StyleDiT), a novel framework that integrates the strengths of StyleGAN with the diffusion model to generate high-quality and diverse kinship faces.
In this framework, the rich facial priors of StyleGAN enable fine-grained attribute control, while our conditional diffusion model is used to sample a StyleGAN latent aligned with the kinship relationship of conditioning images by utilizing the advantage of modeling complex kinship relationship distribution. StyleGAN then handles latent decoding for final face generation.
Additionally, we introduce the Relational Trait Guidance (RTG) mechanism, enabling independent control of influencing conditions, such as each parent's facial image. RTG also enables a fine-grained adjustment between the diversity and fidelity in synthesized faces.
Furthermore, we extend the application to an unexplored domain: predicting a partner's facial images using a child's image and one parent's image within the same framework.
Extensive experiments demonstrate that our StyleDiT outperforms existing methods by striking an excellent balance between generating diverse and high-fidelity kinship faces.
\end{abstract}    
\vspace{-5mm}
\section{Introduction}
\label{sec:intro}

Recent advances in computational facial analysis have significantly improved our understanding of parent-child visual relationships in photographs. This progress is evident in kinship verification \cite{peng2023kfckinshipverificationfair, yu2020deep} and genetic studies \cite{alvergne_differential, cole2017human}, which explore familial facial similarities. Simultaneously, developments in face synthesis and editing have sparked interest in high-fidelity child face generation. This emerging field, focused on creating descendants' faces from parental features, has potential applications in crime investigation, kinship verification, and finding long-lost family member.

Early research on kinship face synthesis \cite{ozkan2018kinshipgan,gao2019will,sinha2020familygan} approached the task as an image-to-image translation problem, learning direct parent-to-child mappings from limited, low-quality kinship data. This often resulted in low-resolution blurry images with little variation.
In contrast, several works \cite{childpredictor, lin2021styledna, kinstyle, stylegene} extracted genetic features from parental face images to generate more faithful child faces.
Recent methods \cite{stylegene, kinstyle, lin2021styledna} leveraged the rich semantic information in the latent space of the pre-trained StyleGAN~\cite{abdal2019image2stylegan, karras2020analyzing, Karras2021} to produce high-quality child faces. StyleGAN's smooth and disentangled latent space allows for seamless fusion of parental features through latent interpolation. However, these approaches still struggle to strike a balance between both high diversity and fidelity in the generated faces while precisely controlling facial attributes. 

Moreover, the control of diversity is essential in kinship face generation since child faces generated from a parent pair can vary widely in resemblance, mirroring the natural variation seen among siblings in a family. Thus, a kinship generation model should produce diverse child faces for a given parental pair while allowing users to control the degree of resemblance to the parents.
Furthermore, an innovative and potentially valuable task of partner face prediction remains largely unexplored. In our empirical study in Appendix \ref{sec:sup_partner_empirical}, a simple linear operation, as used in previous approaches \cite{abdal2019image2stylegan, karras2020analyzing, Karras2021}, between a child and the other parent in the StyleGAN latent space fails to yield satisfactory results. 
This highlights the need for a learning-based approach to tackle this challenging task, as kinship distribution is inherently complex to model.

Inspired by previous works \cite{rishubh2024precisecontrol, li2024stylegan, gandikota2023sliders, dalva2024gantastic} that combine the strengths of StyleGAN and diffusion models, and recognizing the need for precise control over age and gender in kinship tasks, we propose the Style Latent Diffusion Transformer (StyleDiT) to address kinship generation challenges. StyleDiT synergizes the fine-grained and continuous attribute control of StyleGAN's style latent space with the diverse generative capabilities of diffusion models \cite{ddpm,song2020denoising}, leveraging their superior ability to capture the underlying distribution of kinship relationships. In this framework, StyleGAN handles the final face generation, while our conditional diffusion model samples a StyleGAN latent that aligns with the characteristics of the conditioning images.
The proposed StyleDiT excels in generating high-fidelity kinship faces with precise control over attributes like age and gender, while ensuring a diverse range of outputs. Additionally, we introduce the Relational Trait Guidance (RTG) mechanism, an advanced Classifier-Free Guidance~\cite{cfg} technique for kinship face generation. RTG uniquely allows independent control of each influencing condition, such as individual parental facial images. This control ability not only enhances the similarity of the generated results to the input conditions but also enables users to achieve an optimal balance between diversity and fidelity in the images generated based on their specific needs.
Furthermore, we are the first to address the challenging task of predicting a partner's facial features using a child’s image and one parent’s image. This innovative task setting holds significant potential in forensic science for reconstructing missing persons' appearances, as well as in social and genetic research for studying inherited facial traits.
Our main contributions are summarized as follows:
\begin{itemize}
\item We propose StyleDiT, a unified framework for generating diverse and accurate kinship faces with precise control over attributes like age and gender.
\item Our Relational Trait Guidance allows independent control of each influencing factor, enabling users to balance diversity and fidelity in the generated images effectively.
\item Extensive evaluation and user study show that StyleDiT strikes an excellent balance between identity similarity and diversity, outperforming state-of-the-art methods and highlighting its effectiveness and innovation.

\end{itemize}

\begin{figure*}[ht!]
    \centering
    \includegraphics[width=1.0\textwidth]{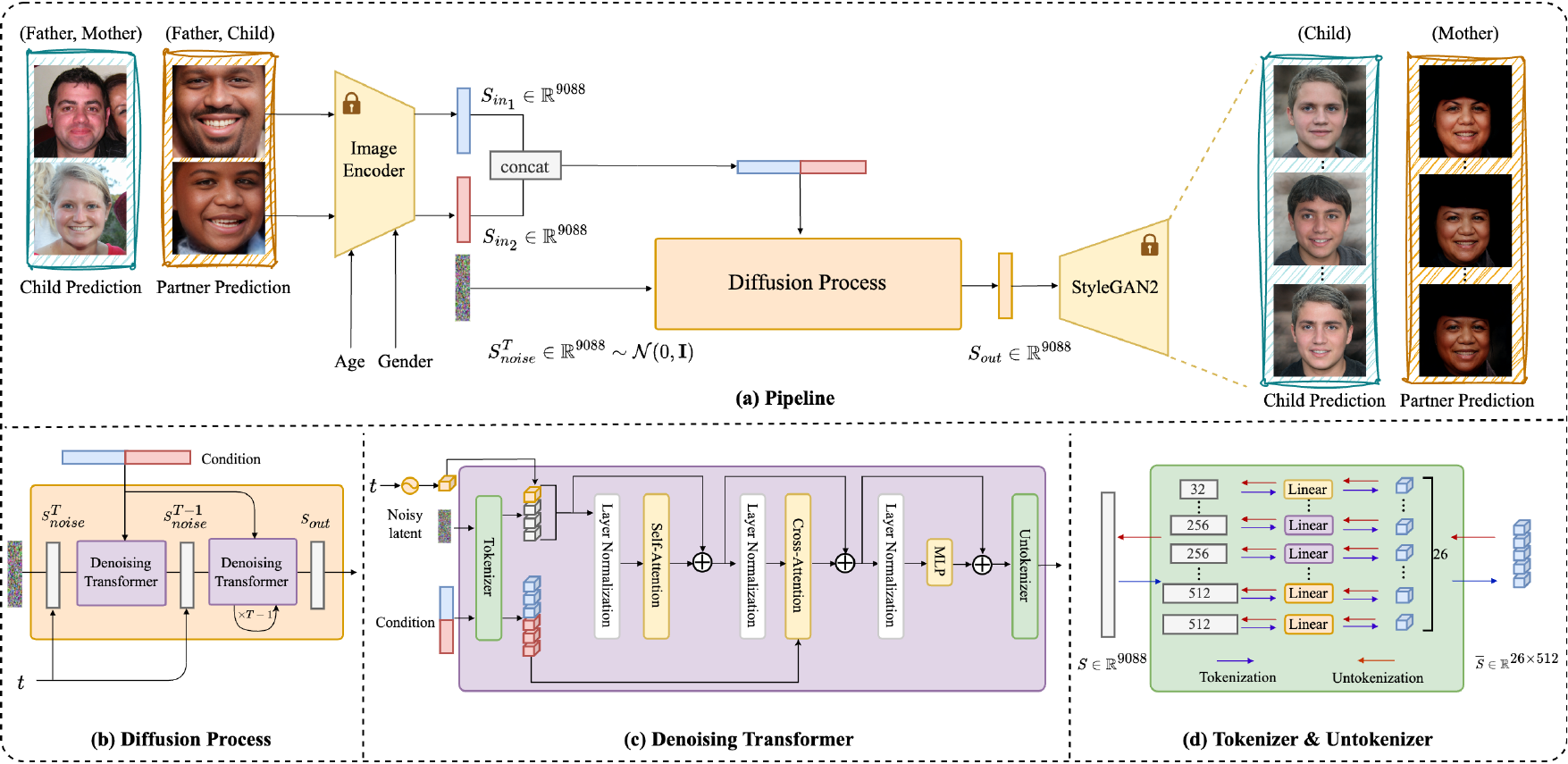}
    \caption{
The overview of the proposed framework.
For both child and partner prediction tasks, input images are first encoded using an image encoder. The encoded style latents, $S_{in_1}$ and $S_{in_2}$, serve as conditions during the diffusion process. The sampled noisy latent undergoes processing through multiple Denoising Transformer blocks, resulting in the predicted face's latent $S_{out}$. Finally, StyleGAN2 decodes this predicted latent to generate a high-fidelity kinship face. The lock icon indicates that the block is frozen during training.}
    \label{fig:pipeline}
    \vspace{-15pt}
\end{figure*}
\section{Related Work}
\label{sec:related}

\paragraph{Deep Image Generation and Manipulation.}
Recent advancements in image generation and manipulation have been driven by innovative generative models. 
Generative Adversarial Networks (GANs) \cite{goodfellow2020generative}, particularly in frameworks like StyleGAN \cite{karras2019style,karras2020analyzing,Karras2021}, enable precise manipulation and adjustment within high-dimensional latent spaces via GAN inversion, facilitating detailed edits of real images \cite{shen2020interfacegan, yao2021latent}. Additionally, diffusion models \cite{ddpm,song2020denoising} have emerged as powerful tools, progressively refining image quality through iterative processes. 
Moreover, a promising trend is emerging in frameworks that combine the strengths of StyleGAN and diffusion models \cite{rishubh2024precisecontrol, li2024stylegan, gandikota2023sliders, dalva2024gantastic}. These approaches synergize the disentangled and fine-grained facial attribute control of StyleGAN's latent space with the versatility of text-driven image generation in diffusion models. Our StyleDiT framework advances this concept by integrating the precise facial attribute manipulation of StyleGAN's latent space with the diffusion model's ability to capture the underlying distribution of kinship relationships. Specifically, we incorporate the $S$ space \cite{wu2021stylespace} of StyleGAN into a one-dimensional transformer-based diffusion model \cite{erkocc2023hyperdiffusion,peebles2023scalable}, moving beyond the conventional focus on predicting latent codes in the $W$ or $W+$ spaces \cite{pinkney_clip2latent,zhang_styleavatar3d}.

\paragraph{Kinship Face Synthesis}
Past works in kinship face synthesis \cite{ozkan2018kinshipgan,gao2019will,sinha2020familygan} usually use supervised learning for parent-to-child mapping. ChildPredictor \cite{childpredictor} introduces a disentangled representation learning approach to separate genetic influences, enhancing parental trait transfer to synthesized faces. However, due to scarcity and the low-quality of kinship data, these methods are prone to overfitting and often produce images with low-quality. In contrast, recent works \cite{stylegene, lin2021styledna, kinstyle} leverage StyleGAN to generate high-resolution child faces by latent interpolation. StyleDNA \cite{lin2021styledna} learns the parent-to-child mapping in ${W}$ space by supervised learning, while KinStyle \cite{kinstyle} employs an image encoder accurately capturing facial traits with precise attribute control and further fine-tunes with real data to try to learn the underlying kinship relationship. However, KinStyle is limited to generating a single outcome per parent pair and often struggles with unnatural skin tones. StyleGene \cite{stylegene} develops a regional facial gene extraction framework with the gene pool to diversify the results, but lacks precise age and gender control. Despite these advances, optimally balancing between diversity and fidelity with fine-grained age and gender manipulation remains challenging for these models.\\
\indent To overcome these challenges, our proposed StyleDiT framework merges the strengths of StyleGAN and diffusion model. 
This approach enables the generation of highly diverse and faithful descendants, with fine-grained and continuous control over age and gender. Additionally, our RTG allows for independent control of each condition and enables users to achieve an optimal balance between diversity and fidelity according to their specific needs.
\section{Method}
To our knowledge, the proposed method is the first unified framework designed for dual capabilities: synthesizing child faces from parental faces and generating partner faces using the child’s face and one parent’s image. A notable feature of our method is its ability to produce diverse results while maintaining high fidelity in both tasks and offering fine-grained and continuous attribute control. 
\subsection{Preliminary}
\noindent\textbf{Child and Partner Face Synthesis.} For the task of predicting the child faces, denoted as $I_{C}$, our method requires a pair of parental face images: the father's face, $I_{F}$, and the mother's face, $I_{M}$. In the task of synthesizing the partner faces, represented as $I_{P_{out}}$, the input consists of the child's image, $I_{C}$, and a partner face image, $I_{P_{in}}$. In this context, $I_{P_{out}}$ and $I_{P_{in}}$ refer to either the father's image ($I_{F}$) or the mother's image ($I_{M}$), depending on the target partner face. Our framework is designed to synthesize kinship faces with high fidelity and diversity for both tasks with a specified age $\alpha$ and gender $\beta$. Formally, the objective of our framework $F$, which predicts a face based on two input images $I_{in_{1}}$ and $I_{in_{2}}$, age $\alpha$ and gender $\beta$, is defined as follows:
\begin{equation}
I_{out} = F(I_{in_{1}}, I_{in_{2}}, \alpha, \beta).
\label{eq:object}
\end{equation}
\noindent\textbf{Diffusion Model.} To understand our architecture, we begin with a primer on diffusion models~\cite{sohl2015deep,ddpm} which in general consist of forward noising and backward denoising processes. 
For the forward noising process, it incrementally adds noise to the initial data $x_0$, formalized in $q(x_t|x_0) = \mathcal{N}(x_t; \sqrt{\bar{\alpha}_t}x_0, (1 - \bar{\alpha}_t)\mathbf{I})$, where $\bar{\alpha}_t$ are key hyperparameters. The sampling of this process is achieved using the reparameterization trick: $x_t = \sqrt{\bar{\alpha}_t}x_0 + \sqrt{1 - \bar{\alpha}_t} \epsilon_t$, with $\epsilon_t$ drawn from a standard normal distribution. In addition, the reverse process of the model, aimed at removing the added noise, is defined as $p_\theta(x_{t-1}|x_t) = \mathcal{N}(\mu_\theta(x_t), \Sigma_\theta(x_t))$, with neural networks predicting $p_\theta$'s parameters. Training is guided by optimizing the variational lower bound on the data's log-likelihood, represented in Eq. \ref{eq:diffusion2}.
\begin{equation}
\mathcal{L}(\theta) = -p(x_0|x_1) + \sum_t \mathcal{D}_{KL}(q^*(x_{t-1}|x_{t},x_0)||p_\theta(x_{t-1}|x_t)).
\label{eq:diffusion2}
\end{equation}
The Kullback-Leibler divergence $\mathcal{D}_{KL}$ between the Gaussian distributions $q^*$ and $p_\theta$ is computed via their means and covariances, while training minimizes Mean Squared Error (MSE), either between the denoised and input data or the predicted and actual noise.

\subsection{Pipeline}
Motivated by previous works \cite{rishubh2024precisecontrol, li2024stylegan} that merges the strengths of StyleGAN and diffusion model, and recognizing the importance of precise control over age and gender in kinship tasks, our proposed framework, illustrated in Fig.~\ref{fig:pipeline}{\color{cvprblue}a}, combines the fine-grained attribute control of StyleGAN's style latent space with the diverse generative capabilities of diffusion models to effectively fitting complex distribution. In our framework, input face images $I_{in_{1}}$ and $I_{in_{2}}$ are first encoded into style latent codes, $S_{in_{1}}$ and $S_{in_{2}}$, using our pre-trained image encoder for precise age and gender control. For more implementation details of our pre-trained image encoder, please refer to Appendix \ref{sec:sup_image-encoder}. We then introduce StyleDiT, which integrates these style latents with a transformer-based conditional diffusion model. By leveraging the diffusion model’s strength in capturing sophisticated kinship relationship distributions, StyleDiT uses $S_{in_{1}}$ and $S_{in_{2}}$ to generate a diverse set of predicted latent codes, $S_{out}$, overcoming the limitations of previous methods \cite{lin2021styledna, kinstyle} that could only produce a single deterministic prediction. Finally, StyleGAN2 decodes $S_{out}$ to produce the final image, $I_{out}$.

\subsection{Style Latent Diffusion Transformer}

\begin{figure*}[t]
    \includegraphics[width=\textwidth]{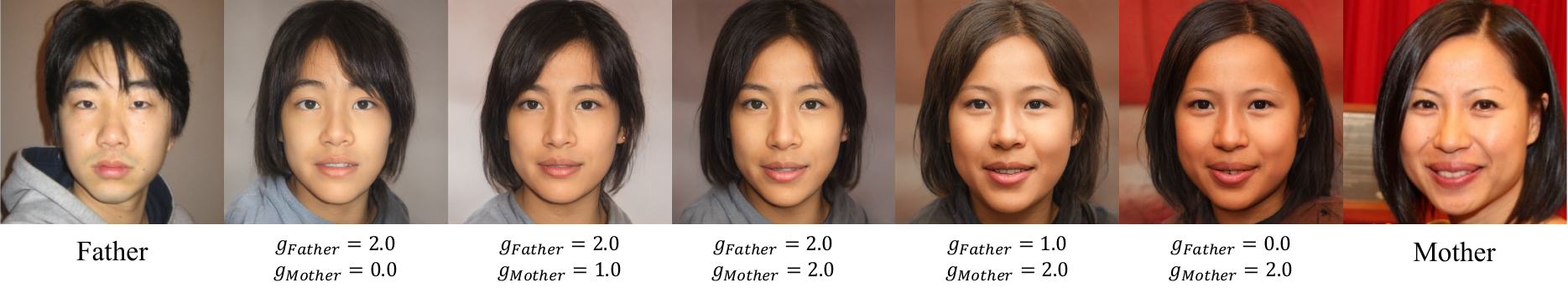}
    
    \vspace{-5pt}
    \caption{The effect of different guidance scales during inference. Progressing from left to right, each image results from varying pairs of guidance scales, with higher scales producing synthesized images that more closely resemble the specified conditions.}
    \label{fig:cfg}
    \vspace{-10pt}
\end{figure*}

We present Style Latent Diffusion Transformer (StyleDiT), an innovative architecture tailored for kinship tasks that efficiently models the style latent space of StyleGAN. 
In our approach, we model the predicted style latent, based on the condition latent, using a diffusion process as depicted in Fig.~\ref{fig:pipeline}{\color{cvprblue}b}. The transformer architecture, denoted as $\mathcal{T}$, functions as denoising network. As shown in Fig.~\ref{fig:pipeline}{\color{cvprblue}c}, the transformer receives an initial arbitrary Gaussian noise $S_{noise}^T \sim \mathcal{N}(0,\mathbf{I})$, alongside condition inputs $S_{in_{1}}$ and $S_{in_{2}}$. It then iteratively denoises $S_{noise}^T$ into $S_{noise}^0=S_{out}$, based on $S_{in_{1}}$ and $S_{in_{2}}$. At each denoising timestep $t$, $S_{noise}^t$, $S_{in_{1}}$, and $S_{in_{2}}$ are fed into a tokenizer (see Fig.~\ref{fig:pipeline}{\color{cvprblue}d}) which categorize style latents into distinct groups, each governing specific facial feature.

According to StyleSpace \cite{wu2021stylespace}, the style latent dimension $S \in \mathbb{R}^{9088}$ represents the total number of style parameters affecting various layers and tRGB blocks in StyleGAN2 \cite{karras2020analyzing}. The style parameters in StyleGAN2, which differ across its various layers, are initially segmented from a set $ S $ in the 9088-dimensional real space ($ \mathbb{R}^{9088} $). Each group of these parameters is then mapped onto tokens with a uniform embedding size of 512 using separate linear layers. Given that there are 26 distinct groups of style parameters, 26 unique linear layers are used for this projection. After the style parameters being tokenized, tokens $\overline{S} \in \mathbb{R}^{26 \times 512}$ are obtained. The sinusoidal embedding of timestep $t$ is transformed into a timestep token via an additional linear projection layer. This token, along with the noisy token $\overline{S}_{noise}^t$, is combined with a learnable positional encoding vector and serves as the input for the transformer.

Conversely, $\overline{S}_{in_{1}}$ and $\overline{S}_{in_{2}}$ are concatenated and summed with the same learnable positional encoding vector as $\overline{S}_{noise}^t$ and act as conditions in the denoising process, guided by a cross-attention mechanism. The transformer outputs denoised tokens $\overline{S}_{noise}^{t-1}  \in \mathbb{R}^{26 \times 512}$, which are then reverted to $S_{noise}^{t-1} \in \mathbb{R}^{9088}$ through an untokenizer comprising another set of linear layers. 
Notably, the transformer $\mathcal{T}$ directly predicts the denoised style latent rather than the noise. We employ Denoising Diffusion Implicit Models (DDIM) \cite{song2020denoising} for our diffusion process.
The training phase involves adding noise into the initial style latent $S$, through a forward diffusion process. We train the network by MSE loss to compare the denoised style latent $S^*$ against the original style latent $S$. Note that the image encoder and StyleGAN2 remain frozen during StyleDiT training. More details for training, please refer to Appendix \ref{sec:sup_training_details}.
During inference, a variety of outcomes are generated by sampling different instances of $S_{noise}^T$ under the consistent conditions of $S_{in_{1}}$ and $S_{in_{2}}$, thus capitalizing on the diffusion process's capability for diverse output generation.

\subsection{Multi-Conditional Classifier-Free Guidance}
Our innovative framework, utilizing a diffusion process, is adept at producing a range of results from a single pair of input images. This versatility addresses a key limitation observed in some prior studies \cite{lin2021styledna, kinstyle}, which were restricted to generating deterministic outcomes based on fixed conditions of input faces, age, and gender. Specifically, these methods could only provide a single prediction, without the capability to sample multiple variations, thereby reducing flexibility in kinship face generation tasks.
To better manage the trade-off between fidelity and diversity, we introduce Relational Trait Guidance (RTG), inspired by Classifier-Free Guidance \cite{cfg, spatext}. RTG enables precise control over each condition’s influence, allowing for independent modulation of the final outcome.

\noindent\textbf{Relational Trait Guidance (RTG).}
With RTG, it allows conditional and unconditional models simultaneously and blending their outputs during inference. Our conditional diffusion model is formulated as $S^*_\mathcal{T}(S_{noise}^t | \{S_{in_{j}}\}_{j=1}^{2})$, following a dual-style latent conditioning setup that allows for precise control over each condition during inference \cite{spatext}. In the training phase, each condition $S_{in_{j}}$ is replaced by the null condition $\emptyset$ with a fixed probability.
During inference, the direction of each condition is identified as $\Delta^t_j = S^*_\mathcal{T}(S_{noise}^t | S_{in_{j}}) - S^*_\mathcal{T}(S_{noise}^t | \emptyset)$, calculated individually. These directions are subsequently integrated using two guidance scales $\{g_j\}_{j=1}^{2}$ as follows:
\begin{equation}
\label{eqn:multiscale_cfg}
    \vspace{-5pt}
    \hat{S^*}_\mathcal{T}(S_{noise}^t | \{S_{in_{j}}\}_{j=1}^{2}) = S^*_\mathcal{T}(S_{noise}^t | \emptyset) + \sum_{j=1}^{2} g_j \Delta^t_j.
\end{equation}
During training, we randomly assign $S_{in_{1}} = \emptyset$ for 10\% data, $S_{in_{2}} = \emptyset$ for 10\% data, and both $S_{in_{1}} = \emptyset$ and $S_{in_{2}} = \emptyset$ for 1\% data. $\emptyset$ symbolizes the learnable style latent corresponding to various age and gender group combinations, as determined from our training data.
Please refer to Appendix \ref{sec:sup_RTG} for detailed design of null conditions. 
\section{Experiments}
\label{sec:formatting}

We first outline our experimental settings, including datasets and baselines in Section \ref{sec:exp-settings}. We then present qualitative results to demonstrate our RTG can synthesize highly diverse results with high-fidelity, compare it with baseline methods in child prediction, and showcase partner face synthesis outcomes in Section \ref{sec:qual-eval}. Next, we provide quantitative results, measuring diversity scores to identify the most diverse approach and calculating identity similarity to determine which method best balances diversity and fidelity. We include a user study to evaluate our method's performance in child and partner face prediction and assess the effectiveness of our attribute control in Section \ref{sec:quan-eval}. Finally, we conduct ablation studies to evaluate the effectiveness of using real data and assess the impact of the diffusion process and RTG in Section \ref{sec:ablation}. For more implementation details, please refer to Appendix \ref{sec:sup_implementation}.

\subsection{Experimental Settings}
\label{sec:exp-settings}

\noindent\textbf{Datasets.} 
To train the diffusion component of our framework, we create a simulated dataset to overcome the limitations of real kinship datasets, including issues with resolution, quality, quantity, and diversity. Our approach is inspired by prior studies \cite{kinstyle, stylegene} suggesting that a child's facial features can be inferred through linear interpolation of parental traits. Additionally, medical research \cite{debruine_social, alvergne_differential} supports the idea that offspring inherit genes through random genetic combinations, reinforcing the concept that a child's facial features can be deduced through the interpolation of parental traits. The dataset includes 100,000 synthetic triplets of father, mother, and child, generated using the process described below. We randomly select 70,000 male and female pairs from the CelebA-HQ dataset as parental images, map them into StyleGAN2’s latent space, and apply 200 combinations of age and gender attributes  (0-99 years old $\times$ 2 genders). StyleGAN2 then generates child faces through linear interpolation between the parental style latents, employing variable weights to control the resemblance to either the mother or father, thus ensuring diversity in the resulting child images. Moreover, we sample 30,000 male and female faces directly from StyleGAN2's latent space and follow the same generation procedure as with CelebA-HQ. This simulated dataset is used for training both child and partner face prediction.\\
\indent For evaluation, we use the test splits of the FIW \cite{fiw}, TSKinFace \cite{TSKinFace}, and FF-Database \cite{childpredictor} datasets. Preprocessing involved facial alignment and quality enhancement using CodeFormer \cite{zhou2022towards}, followed by resizing all images to 256 × 256 pixels. Additionally, since the FIW and TSKinFace datasets lack predefined splits and the FF-Database includes only training and validation sets, we conducted custom dataset splits: 296 test triplets for FIW, 149 for TSKinFace, and 368 for FF-Database.\\

\noindent\textbf{Baselines.} Our method is benchmarked against state-of-the-art baselines in the child face synthesis task, including ChildPredictor \cite{childpredictor}, StyleGene \cite{stylegene}, and KinStyle \cite{kinstyle}. We replicate these models' results by adhering to the settings in their original source code and the implementation details described in their respective papers. This allows for a comprehensive comparison of our approach in both quantitative and qualitative assessments. For the partner face synthesis task, we found no existing studies formally addressing this problem at the time of our submission. Therefore, we focus on showcasing the visual quality of the faces generated by our framework for partner face synthesis and provide the corresponding quantitative scores.

\subsection{Qualitative Evaluation}
\label{sec:qual-eval}

\begin{figure}[t]
    \centering
    \includegraphics[width=0.7\columnwidth]{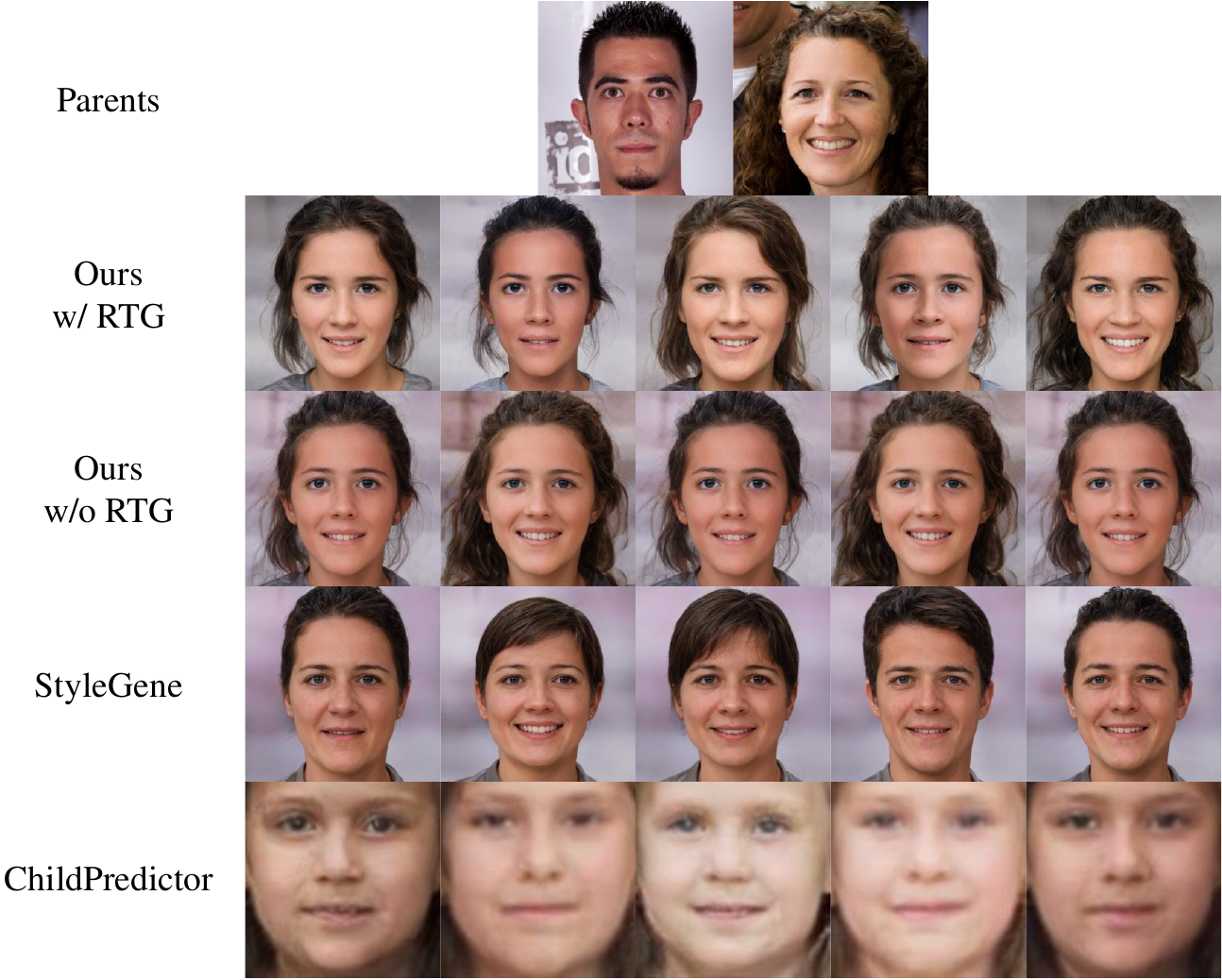}
    \caption{This figure compares the facial diversity in results of our method, StyleGene \cite{stylegene} and ChildPredictor \cite{childpredictor}. All faces are generated with specified attributes of 15 years old and female.}
    \label{fig:diversity}
    \vspace{-20pt}
\end{figure}

\noindent\textbf{Effectiveness of Relational Trait Guidance.} Our innovative Relational Trait Guidance technique offers the flexibility to prioritize either diversity or fidelity in the generated images. With RTG, our method produces diverse child prediction results shown in Fig.~\ref{fig:diversity}. In comparison, our method without RTG produces high-fidelity but less varied results. StyleGene \cite{stylegene} generates varied faces but lacks the capability to precisely manipulate age and gender attributes, particularly when the objective is to maintain higher resemblance to parents. KinStyle \cite{kinstyle}, while generating high-fidelity kinship faces, offers limited diversity, producing only a single prediction result. Finally, ChildPredictor~\cite{childpredictor} generates diverse child faces with lower-quality images compared to other approaches. By employing RTG, as depicted in Fig.~\ref{fig:cfg}, we can selectively generate child faces to resemble either the father or mother by adjusting their respective guidance scales during inference for child prediction.

\begin{figure*}[t]
    \begin{subfigure}{0.5\textwidth}
        \centering
        \includegraphics[width=\linewidth]{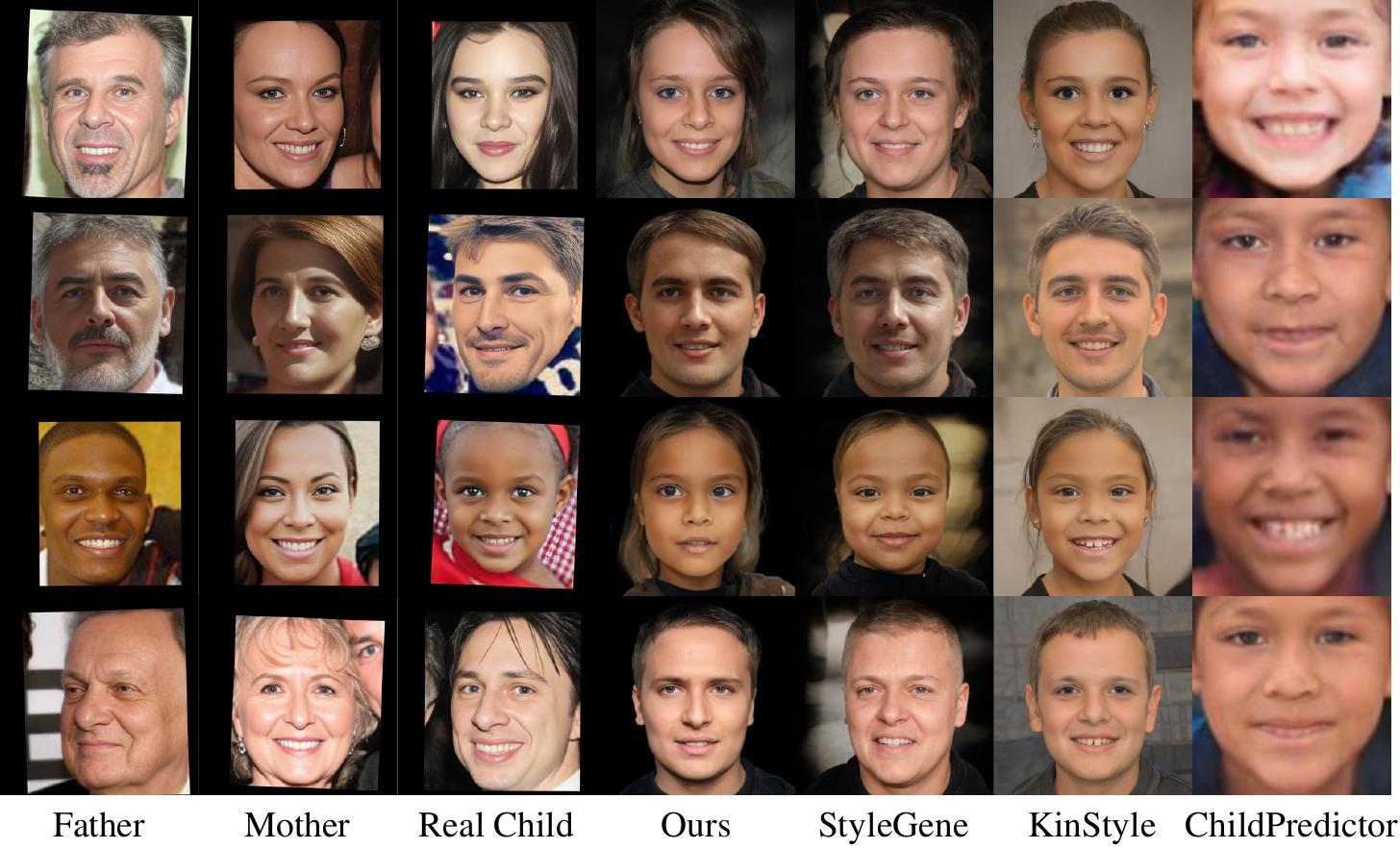}
        \caption{FIW dataset}
        \label{fig:subfig1}
    \end{subfigure}%
    \begin{subfigure}{0.5\textwidth}
        \centering
        \includegraphics[width=\linewidth]{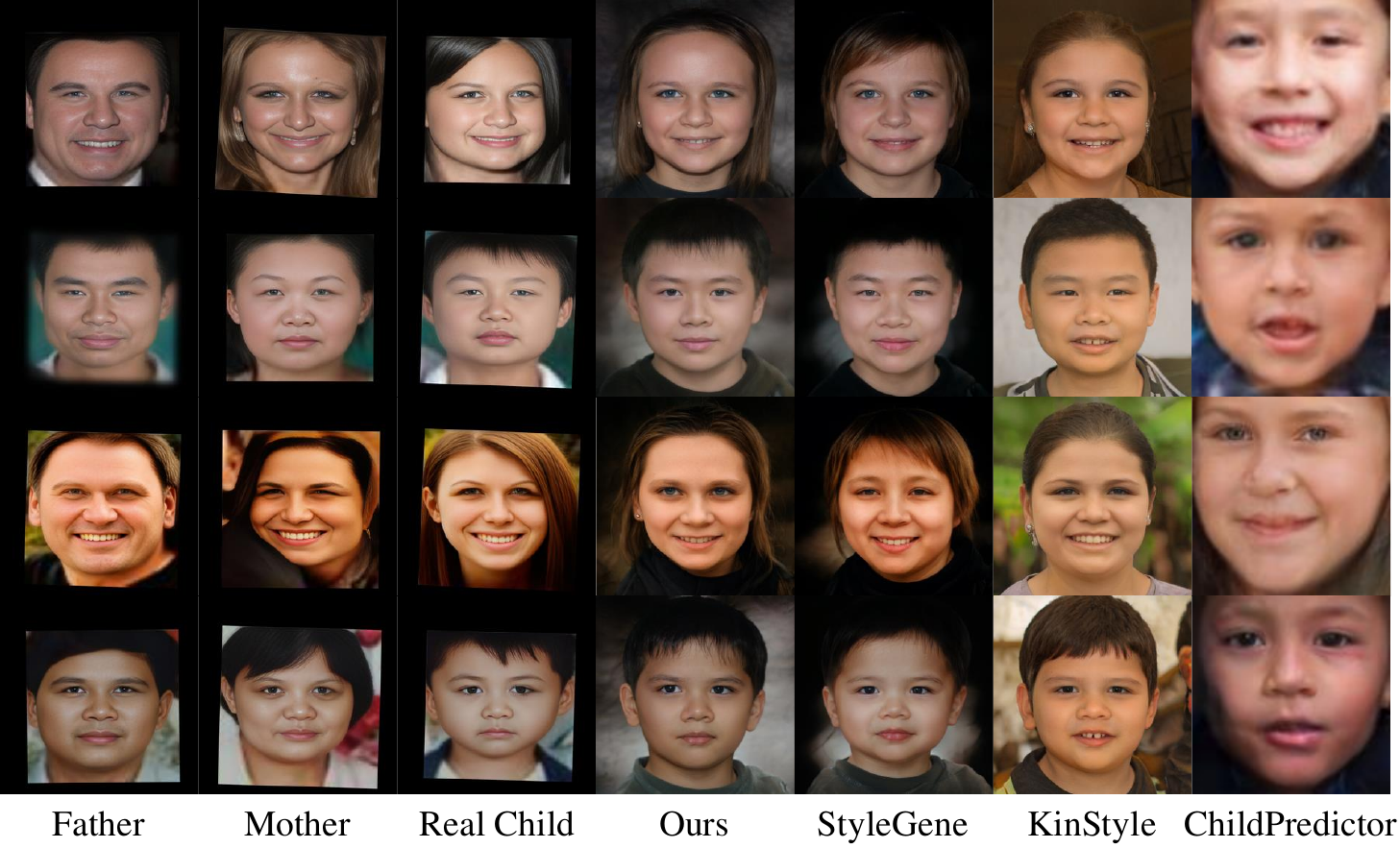}
        \caption{TSKinFace dataset}
        \label{fig:subfig2}
    \end{subfigure}
    \caption{The qualitative comparison of synthesized children's faces by StyleDiT and baselines. The first three columns in both (a) FIW and (b) TSKinFace datasets display the father, mother, and real children, followed by four different methods of synthesized child images. The results include faces with different races and skin tones to demonstrate the framework's ability to generalize across various racial backgrounds. Each method generates faces according to the gender and age of the real child in each family.}
    \label{fig:baseline-child}
    \vspace{-5pt}
\end{figure*}

\noindent\textbf{Comparison with the State-of-the-Art.}
Fig.~\ref{fig:baseline-child} presents more qualitative results for child prediction task on the FIW and TSKinFace datasets. We conduct a comparative analysis of our approach with StyleGene, KinStyle, and ChildPredictor on these kinship datasets. ChildPredictor shows limitations in generating high-quality images, primarily due to the absence of utilizing the StyleGAN generator. While KinStyle can produce high-quality faces with well-preserved parental identity similarity, its deterministic framework lacks generation diversity and often generates faces with inaccurate skin tones. StyleGene excels at creating diverse, high-quality child face images but struggles to balance the fidelity to parental features with accurate age and gender representation within its gene pool framework. In contrast, our method achieves an exceptional balance in generating diverse, high-quality images that maintain strong parental traits across various racial backgrounds while also offering fine-grained control over age and gender.

\begin{figure}[t]
    \centering
    \includegraphics[width=\columnwidth]{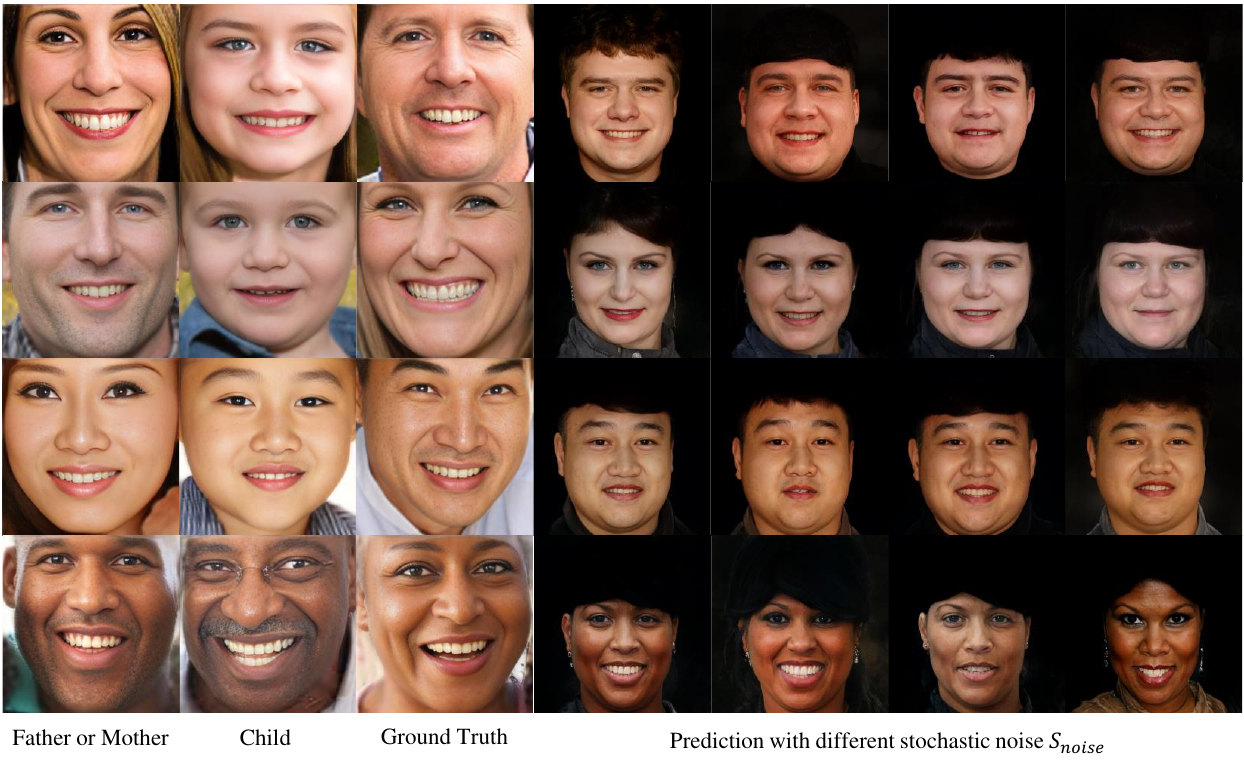}
    \caption{The results demonstrate our framework can synthesize high-fidelity and diverse partner faces across various races.}
    \label{fig:partner-face}
    \vspace{-15pt}
\end{figure}

\noindent\textbf{Partner Face Synthesis.}
In our framework, we extend its functionality beyond synthesizing child face images to generate high-fidelity partner faces. By training on the same datasets and utilizing the same model structure, the framework can synthesize partner faces given a child image and an image of the other partner. Additionally, it showcases the ability to produce diverse synthetic results across various races, as illustrated in Fig.~\ref{fig:partner-face}. For more results and details, please refer to Appendix \ref{sec:sup_partner}.

\subsection{Quantitative Evaluation}
\label{sec:quan-eval}

For our quantitative evaluation, we use the test splits from the FIW, TSKinFace, and FF-Database datasets. For each family, we generate 20 children for the child prediction task and 20 partners for the partner prediction task. These generated images are used to calculate Identity Similarity and measure diversity. We present the result of child prediction task in the below sections, while the partner prediction task results are detailed in the Appendix \ref{sec:sup_partner_quantitative}.

\begin{table}[t]
    \centering
    \resizebox{\columnwidth}{!}{%
    \setlength{\tabcolsep}{2pt}
    \begin{tabular}{lccccccc}
        \toprule
        \multirow{2}{*}{Methods} & \multicolumn{3}{c}{DS ($\downarrow$)} & \multicolumn{3}{c}{ID Sim ($\uparrow$)} \\
        \cmidrule(lr){2-4} \cmidrule(lr){5-7}
        & FIW & TSKinFace & FF-Database & FIW & TSKinFace & FF-Database \\
        \midrule
        ChildPredictor & 0.7805 & 0.7811 & 0.7702 & 0.4996 & 0.4989 & 0.4979 \\
        KinStyle       & --- & --- & --- & \textbf{0.7707} & \textbf{0.7768} & \textbf{0.7772} \\
        StyleGene      & \underline{0.6757} & \underline{0.7496} & \textbf{0.6966} & \underline{0.7132} & 0.7242 & 0.6887 \\
        StyleDiT (Ours) & \textbf{0.6140} & \textbf{0.6780} & \underline{0.7078} & 0.7003 & \underline{0.7244} & \underline{0.7124} \\
        \bottomrule
    \end{tabular}%
    }
    \caption{Quantitative results of diversity (DS) and identity similarity (ID Sim) between predicted child and ground truth child. Our method strikes an excellent balance between diverse synthesis and reasonably high fidelity, although it does not surpass KinStyle in identity similarity, as it can only generate deterministic results.}
    \label{table:idsim-and-diversity}
    \vspace{-10pt}
\end{table}

\noindent\textbf{Diversity Measurement.}
We measure diversity by calculating the pairwise cosine similarity of ArcFace \cite{arcface} features from various predicted outcomes. These outcomes are derived by sampling with different random seeds or specific parameters, yet all originate from the same pair of parent images.  
While these results exhibit diversity, they are limited by the requirement to maintain consistent age and gender across all outcomes.
For diversity assessment, each image is first transformed into a facial feature using ArcFace, pre-trained on the MS-Celeb-1M \cite{msceleb} dataset, to capture critical semantic and facial details. The diversity score $DS$ is calculated as per Eq. \ref{eq:diversity}, considering $N$ as the number of samples and $x_i$, $x_j$ as individual feature vectors.
\begin{equation}
DS = \frac{1}{N(N-1)} \sum_{i \neq j} \frac{x_i \cdot x_j}{\|x_i\|\|x_j\|}
\label{eq:diversity}
\end{equation}
A lower mean similarity score signifies greater diversity, implying that the generated images, despite sharing the same input condition, exhibit less resemblance to each other when sampled with varying parameters like random seeds.

In Table \ref{table:idsim-and-diversity}, we compare the $DS$ score of StyleGene, ChildPredictor, and our method. We exclude KinStyle due to its single-result limitation. Our method shows the lowest DS in the FIW and TSKinFace datasets, and the second lowest in FF-Database, highlighting its superior ability to generate diverse images over other state-of-the-art methods.

\noindent\textbf{Identity Similarity.}
To evaluate the effectiveness of our child and partner face synthesis methods, we employ the Area Under the Receiver Operating Characteristic (AUC-ROC) curve as the principal metric. This curve measures the model's ability to produce high-fidelity child and partner faces by plotting the true positive rate against the false positive rate at various discrimination thresholds. For a more accurate assessment, we utilize ArcFace to extract facial features before computing the AUC-ROC. This step ensures a comprehensive and robust evaluation of our model's performance in synthesizing both child and partner faces.

We generate 20 faces per family from the test split of each dataset for each method, creating positive and negative pairs by comparing images of the same person for positive pairs and images of different individuals for negative pairs. The results, shown in Table \ref{table:idsim-and-diversity}, reveal that while our method does not surpass KinStyle in identity similarity for child prediction task, it achieves a superior balance between generating diverse outputs and maintaining high fidelity. In contrast, KinStyle can only generate deterministic results.

\begin{table}
  \setlength{\tabcolsep}{3pt}
  \resizebox{1\linewidth}{!}{  
  \begin{tabular}{ccccc}
    \toprule
     \space & Age MSE ($\downarrow$) & Gender ACC ($\uparrow$) & ID Sim ($\uparrow$) \\
    \midrule
    StyleDiT (Ours) & \textbf{0.0023} & \textbf{0.9990} & \textbf{0.7032} \\
    StyleGene~$^\dagger$ & 0.0146 & 0.7315 & 0.6887 \\
    StyleGene~$^\ddagger$ & 0.0047 & 0.9090 & 0.6985 \\
    \bottomrule
  \end{tabular}}
  \caption{Comparison of attribute control. ID Sim represents the identity similarity between predicted child and real child. StyleGene~$^\dagger$ denotes 10\% utilization of the gene pool in the linear combination of parents and the gene pool, while StyleGene~$^\ddagger$ indicates 90\% usage. StyleDiT uses 2.0 for both parental guidance.}
  \label{table:age-gender}
  \vspace{-10pt}
\end{table}

\noindent\textbf{Effectiveness of Fine-grained Age and Gender Control.}
In our framework, StyleGAN's style latent space offer disentangled and fine-grained attribute control. Similarly, StyleGene also generate diverse kinship faces with controlled attributes such as age and gender. However, their attribute control relies on a varied gene pool encompassing multiple ages, genders, and races. This reliance can potentially compromise the fidelity of the input images, particularly when a significant amount of gene pool information is required for precise attribute control.

To evaluate our framework's ability to balance attribute control with image fidelity, we conducted a comparative analysis using 100 randomly selected families from the FIW dataset. For each family, we generated 20 child images. The analysis, as illustrated in Table \ref{table:age-gender}, included assessing the MSE for age and the accuracy of gender predictions. We also measured the preservation of input characteristics by calculating the ID Sim between the predicted child and the real child. Our results indicate that, unlike other approaches, our framework effectively balances attribute control with high fidelity. However, the findings also highlight an inherent trade-off in StyleGene between precise attribute control and fidelity, which is influenced by the extent of the gene pool used. For qualitative results demonstrating precise age and gender control, please refer to Appendix \ref{sec:sup_attribute_control}.

\noindent\textbf{User Study.}
Building on the methodology outlined in StyleGene, we undertake a comparable user study with 100 participants  to assess the effectiveness of our approach. 
Child faces generated by StyleDiT, StyleGene \cite{stylegene}, KinStyle \cite{kinstyle}, and ChildPredictor \cite{childpredictor}, based on 7 parents, are presented to each evaluator who ranks the synthetic child of different methods in terms of similarity with parents. Additionally, a separate assessment focuses on the similarity between predicted and real children, where three children are generated based on three other parents, and evaluators then rank the methods according to identity similarity. 
Combining the results from both parts, as shown in Table \ref{table:user-study}, our method received the highest number of votes among all first-place votes, totaling 1,000 votes, surpassing KinStyle and StyleGene by 172 votes and 47 votes, respectively. Additionally, our method achieved the highest average ranking, demonstrating that our model excels in child face synthesis compared to other baselines.
Moreover, we extend the study to partner face synthesis within our framework where the participants are asked to give a score from 1 to 5 (from most dissimilar to most similar). 
For more details and results, please refer to Appendix \ref{sec:sup_user_study}.

\begin{table}[t]
    \resizebox{\columnwidth}{!}{%
    \setlength{\tabcolsep}{2pt}
    \begin{tabular}{@{}ccccc@{}}
        \toprule
         & ChildPredictor & StyleGene & KinStyle & StyleDiT (Ours) \\
        \midrule
        \# of 1st Votes of Child (\(\uparrow\)) & 76 & 334 & 209 & \textbf{381} \\
        Child Avg. Rank (\(\downarrow\)) & 3.55 & 2.09 & 2.38 & \textbf{1.99} \\
        \bottomrule
    \end{tabular}%
    }
    \caption{The number of 1st votes and average rank of different approaches for child prediction task
    in the user study.}
    \vspace{-15pt}
    \label{table:user-study}
\end{table}

\subsection{Ablation Study}
\label{sec:ablation}

In this section, we conduct ablation studies to evaluate the effectiveness of incorporating real data and assess the impact of the diffusion process and RTG.
\begin{table}[t]
    \centering
    \begin{tabular}{cccc}
         \toprule
            Methods        & FIW   & TSKinFace & FF-Database \\
            \midrule
            StyleDiT (Ours)  & \textbf{0.7003} & \textbf{0.7244} & \textbf{0.7124} \\
            StyleDiT~$^\dagger$ & 0.6817 & 0.6198 & 0.6398 \\
            StyleDiT~$^\ddagger$ & 0.5362 & 0.5181 & 0.5261 \\
        \bottomrule
    \end{tabular}
    \caption{Performance comparison of identity similarity between our configuration, fine-tuning, and training solely on the real kinship dataset. StyleDiT~$^\dagger$ indicates the fine-tuned version, and StyleDiT~$^\ddagger$ denotes training solely on the real kinship dataset.}
    \label{table:quant-real-data}
\end{table}

\noindent\textbf{Effectiveness of Using Real Data.}
In our experiments, we did not observe overall performance improvements when fine-tuning our model or training solely on the real kinship dataset. As shown in Table \ref{table:quant-real-data}, both fine-tuning under our configuration and training from scratch on the kinship dataset led to reduced identity similarity. We hypothesize that this is due to the low quality and scarcity of real kinship data, even with the application of super-resolution techniques \cite{zhou2022towards}, which limits the model's ability to effectively learn complex kinship distributions. Consequently, all results presented in this paper are based on models trained solely on simulated data, which we determined to be the optimal setting. The FF-Database was used as the real kinship dataset in Table \ref{table:quant-real-data}. For additional results on other kinship datasets and further discussion on fine-tuning with real data, please refer to Appendix \ref{sec:sup_real_data}.

\noindent\textbf{Effectiveness of Diffusion Process and RTG.}
The assessment includes identity similarity (ID Sim) and diversity (DS), computed through the test split of the FIW dataset. Table \ref{table:ablation} demonstrates that integrating the diffusion process leads to improvements in DS for the synthesized descendants. As expected, the use of RTG effectively enhances the diversity of synthesized faces with a favorable ID Sim.

\begin{table}[t]
  \centering
  \begin{tabular}{@{}cccccc@{}}
    \toprule
     & Transformer & Diffusion & RTG & ID Sim ($\protect\uparrow$) & DS ($\protect\downarrow$) \\
    \midrule
          & \checkmark & \checkmark & \checkmark & 0.7003 & \textbf{0.6140} \\
          & \checkmark & \checkmark & \space & 0.7636 & 0.9984 \\
          & \checkmark & \space & \space & 0.7666 & --- \\
    \bottomrule
  \end{tabular}
  \caption{Ablation study on FIW dataset.}
  \label{table:ablation}
\end{table}
\section{Conclusion}
We present StyleDiT, a unified framework for generating diverse and high-fidelity kinship faces with precise age and gender manipulation. Our Relational Trait Guidance enables independent control of each influencing factor, allowing users to effectively balance diversity and fidelity. Extensive benchmark evaluations and user studies show that StyleDiT outperforms state-of-the-art methods, striking an excellent balance between identity similarity and diversity, underscoring its effectiveness and innovation.
{
    \small
    \bibliographystyle{ieeenat_fullname}
    \bibliography{main}
}

\appendix
\setcounter{table}{0}
\setcounter{figure}{0}    
\clearpage
\setcounter{page}{1}
\maketitlesupplementary

\begin{figure*}[h!]
    \centering
    \includegraphics[width=1.0\textwidth]{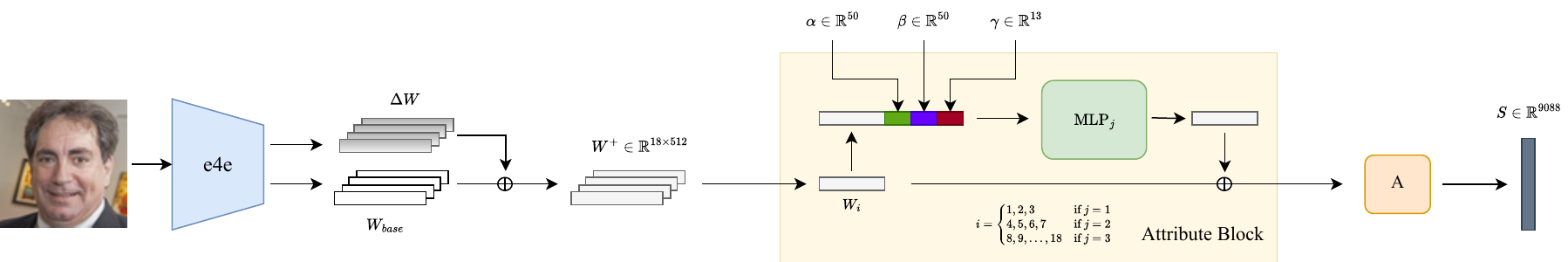}
    \caption{The overview of our image encoder. The input image is first encoded into latent code using e4e. Then, the latent code is adjusted for age, gender, and pose attributes using an attribute block, and finally projected into the $S$ space through affine transformer layers.}
    \label{fig:image_encoder}
\end{figure*}

\begin{figure*}
    \centering
    \includegraphics[width=1.0\textwidth]{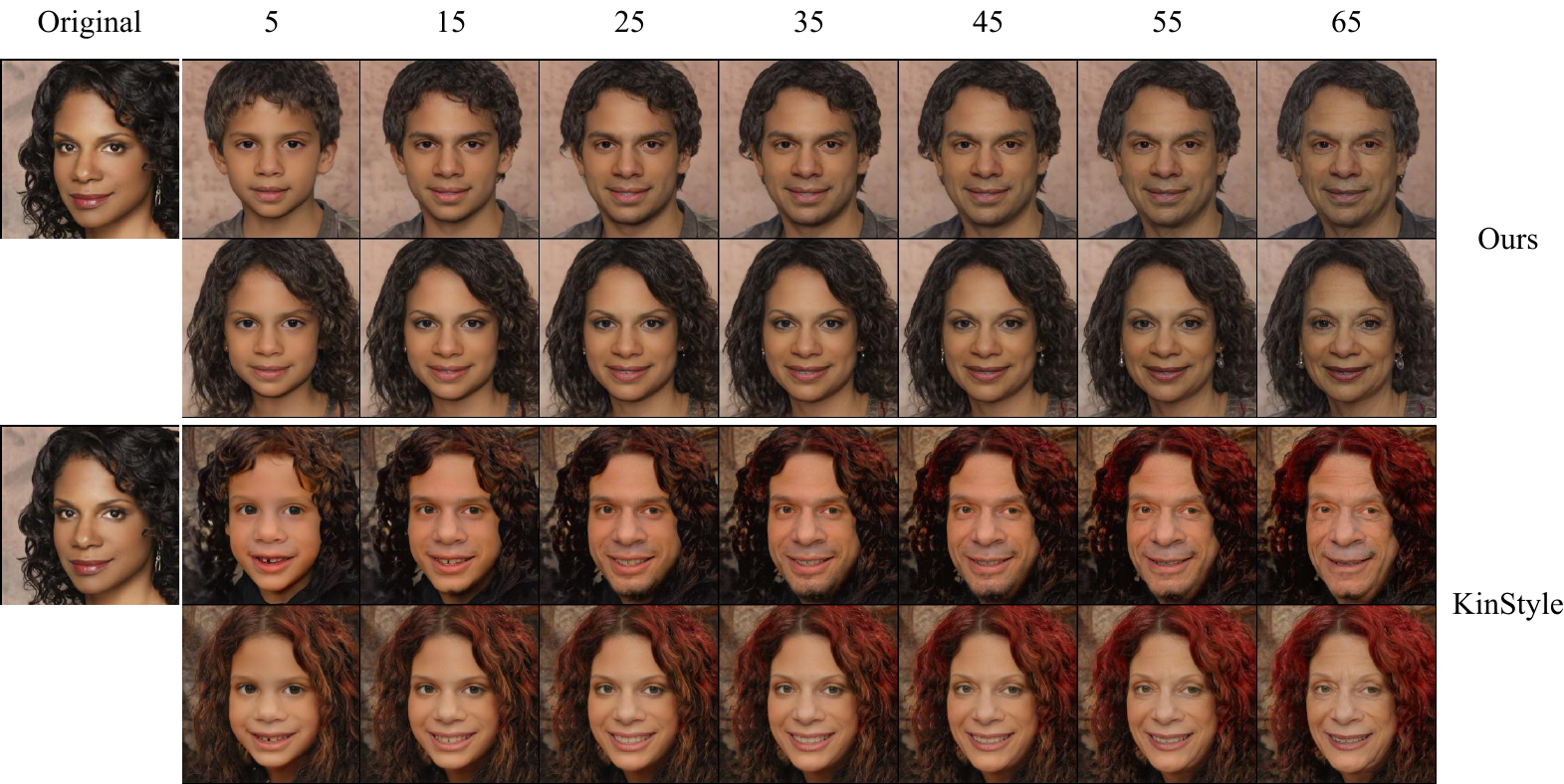}
    \caption{The comparison of controlling age, gender, and frontalization while preserving skin tone. For the first and second rows, our image encoder was used, and for the third and fourth rows, the Kinstyle image encoder was used, with all images generated by StyleGAN2.}
    \label{fig:age-pose_control}
\end{figure*}

\section{Image Encoder}
\label{sec:sup_image-encoder}
Inspired by KinStyle's \cite{kinstyle} effective control over age and gender attributes, as well as its ability to accurately frontalize faces, we adopt its framework as the foundation for our image encoder. However, KinStyle often produces inaccurate skin tones. To address this, we integrate a pre-trained StyleGAN encoder \cite{tov2021designing} and an Attribute Block, which not only controls age and gender but also frontalizes the face simultaneously. This approach preserves the identity, aligns age and gender, frontalizes the face, and maintains the original skin tone, enabling more effective facial attribute manipulation. Figure~\ref{fig:age-pose_control} illustrates a comparison between KinStyle's image encoder and our own in controlling pose, age, and gender. The framework of our image encoder is shown in Figure~\ref{fig:image_encoder}. The design and implementation of each component are detailed in the following sections.

\subsection{StyleGAN Encoder} 

We employ a pre-trained Encoder for Editing (e4e) to map input images to the $W{+}$ space in StyleGAN2 \cite{karras2020analyzing}, ensuring the preservation of identity. This encoder generates latent codes by combining a base latent and a residual, both represented in $ \mathbb{R}^{18 \times 512} $, which enhances the editability and consistency of the resulting latent codes, making them more suitable for fine-grained facial manipulations.

\subsection{Attribute Block} 
The Attribute Block is further designed to align age, gender, and pose of input latent codes in the $W{+}$ space by adjusting latent codes with an offset vector. This vector is learned using MLPs with leaky ReLU, based on several inputs: the concatenated latent codes, a 0-1 age scale $\alpha$ (where 0 represents infancy and 1 represents 100 years old), binary gender values $\beta$ and pose parameters $\gamma$. For pose conditioning, the yaw angle is divided into 13 dimensions, each representing a 15$^\circ$ interval from -90$^\circ$ to 90$^\circ$. To generate a more continuous pose transformation, we interpolate between two neighboring pose dimensions. For example, a 7.5$^\circ$ pose, located between 0$^\circ$ and 15$^\circ$, is represented by interpolating as [0, 0, 0, 0, 0, 0, 0.5, 0.5, 0, 0, 0, 0, 0].\\
\indent During training, we utilize the FFHQ-Aging dataset \cite{karras2019style}, which provides images annotated with age and gender labels. For pose annotation, we employ FaceXFormer \cite{facexformer}, a pre-trained model capable of a wide range of facial analysis tasks. To mitigate data imbalance in age, gender, and pose categories, we implement weighted sampling based on the number of training instances in each category. The block creates three latent code variations—synthesized, reconstructed, and cycle-consistent—derived from the actual age, gender, and pose data. Denoted by $M$, these MLPs operate as follows:
\begin{equation}
\begin{split}
    w'_{syn} &= w + M(w, \alpha, \beta, \gamma)\\
    w'_{rec} &= w + M(w, \alpha_{gt}, \beta_{gt}, \gamma_{gt}) \\
    w'_{cyc} &= w'_{syn} + M(w'_{syn}, \alpha_{gt}, \beta_{gt}, \gamma_{gt})
\end{split}
\end{equation}
These latent codes are then used to produce corresponding images $I_{syn}$, $I_{rec}$, and $I_{cyc}$. The loss function consists of several components. Age, gender, and pose losses are computed using pre-trained classifiers and cross-entropy, where $C_a(\cdot)$, $C_b(\cdot)$, and $C_c(\cdot)$ are the classifiers for age, gender, and pose, and $H(\cdot)$ represents cross-entropy loss:
\begin{equation}
\begin{split}
    \mathcal{L}_{age} = &~||\alpha-C_a(I_{syn})||_2 + ||\alpha_{gt}-C_a(I_{rec})||_2 + \\ &||\alpha_{gt}-C_a(I_{cyc})||_2 \\
    \mathcal{L}_{gen} = &H(\beta,C_b(I_{syn})) + H(\beta_{gt},C_b(I_{rec})) + \\ &H(\beta_{gt},C_b(I_{cyc}))\\
    \mathcal{L}_{pose} = &~||\gamma-C_c(I_{syn})||_2 + ||\gamma_{gt}-C_c(I_{rec})||_2 + \\ &||\gamma_{gt}-C_c(I_{cyc})||_2~,
\end{split}
\end{equation}
The identity loss $\mathcal{L}_{ID}$ ensures the person’s identity is preserved across transformations, where $\xi$ prevents identity degradation:
\begin{equation}
\begin{split}
    \mathcal{L}_{ID} =  &~\xi \cdot (1-<R(I_{syn}),R(I)>) + \\
    &(1-<R(I_{rec}),R(I)>) + \\
    &(1-<R(I_{cyc}),R(I)>),  \\
\end{split}
\end{equation}
We also apply perceptual similarity loss using a pre-trained AlexNet feature extractor $P(\cdot)$ trained on the ImageNet dataset. The term $I_{enc}$ refers to the reconstructed image obtained by passing the latent codes through e4e and then through StyleGAN2. Additionally, $L_2$ regularization is used on the latent codes to ensure they remain within the $W{+}$ space.
\begin{equation}
\begin{split}
    \mathcal{L}_{reg} = &||M(w, \alpha, \beta, \gamma)||_2 + ||M(w, \alpha_{gt}, \beta_{gt}, \gamma_{gt})||_2 + \\ &||M(w'_{syn}, \alpha_{gt}, \beta_{gt}, \gamma_{gt})||_2, \\
    \mathcal{L}_{per} = &||P(I_{enc})-P(I_{syn})||_2 +||P(I_{enc})-P(I_{rec})||_2 +\\  &||P(I_{enc})-P(I_{cyc})||_2 ~,
\end{split}
\end{equation}
These losses are combined into a total loss function, with $\lambda_{ID}$ = 0.5, $\lambda_{age}$ = 8, $\lambda_{gen}$ = 1, $\lambda_{pose}$ = 8, $\lambda_{reg}$ = 0.05, and $\lambda_{per}$ = 0.5.
\begin{equation}
\begin{split}
    \mathcal{L}_{attr} = &\lambda_{ID}\mathcal{L}_{ID} + \lambda_{age}\mathcal{L}_{age} + \lambda_{gen}\mathcal{L}_{gen} + \\ 
    &\lambda_{pose}\mathcal{L}_{pose} + \lambda_{reg}\mathcal{L}_{reg}+\lambda_{per}\mathcal{L}_{per}~,
\end{split}
\end{equation}

Finally, the paper follows the approach suggested by Kafri et al. \cite{stylefusion}, applying affine transform layers from StyleGAN2 to convert each input’s adjusted latent code into the $S$ space to serve as conditional inputs for our diffusion transformer.

\begin{table}
  \resizebox{1\linewidth}{!}{  
  \begin{tabular}{ccccc}
    \toprule
    \multicolumn{5}{c}{Age (year-old)/Gender} \\
    \midrule
     0-2/male & 3-9/male & 10-19/male & 20-29/male & 30-39/male \\
     40-49/male & 50-59/male & 60-69/male & 70-99/male \\
    \midrule
     0-2/female & 3-9/female & 10-19/female & 20-29/female & 30-39/female \\
     40-49/female & 50-59/female & 60-69/female & 70-99/female \\
    \bottomrule
  \end{tabular}}
  \caption{Age and gender group combinations. The table enumerates 18 group combinations of age and gender, covering 9 age categories for each of the two genders.}
  \label{table:age-gender-combinations}
  \vspace{-10pt}
\end{table}

\section{Relational Trait Guidance (RTG)}
\label{sec:sup_RTG}
In this section, we introduce two types of style latents: mean style latents and learned style latents, which serve as null conditions, in Sections \ref{sec:mean-style-latent} and \ref{sec:learned-style-latent}, respectively. Finally, we compare the performance of these two types of style latents in Section \ref{sec:null-performance-comparison} to determine which offers the best results.

\subsection{Mean Style Latent}
\label{sec:mean-style-latent}
Our mean style latents represent 18 distinct age and gender group combinations, as shown in Table \ref{table:age-gender-combinations}. During training, these style latents are aligned with the target age $\alpha$ and gender $\beta$ using the Attribute Block. To generate the mean style latents, we start by sampling 10,000 latent codes ${z}$ from the input latent space ${Z}$ of StyleGAN2. The mapping network ${f
\rightarrow W}$ in StyleGAN2 then produces latent vectors ${w \in \mathbb{R}^{18 \times 512}}$. These ${w}$ latents are transformed to match specific age and gender groups through the Attribute Block in our Image Encoder. Finally, the ${w}$ latents are converted into style latents ${s \in \mathbb{R}^{9088}}$ using affine transformation layers, and the mean style latents for each age and gender combination are obtained by averaging the 10,000 resulting style latents.

\subsection{Learned Style Latent}
\label{sec:learned-style-latent}
During training, we initialize a learnable embedding, ${p \in \mathbb{R}^{1 \times 18 \times 512}}$, as the null condition. To preserve information about age and gender, we pass this embedding through the Attribute Block. Additionally, we randomly assign condition inputs as the null condition with a fixed probability during each training step. In the inference stage, we directly employ the learned embedding as the null condition in each denoising step. 

\subsection{Performance Comparison of Null Conditions}
\label{sec:null-performance-comparison}

\begin{table}
  \resizebox{1\linewidth}{!}{  
  \begin{tabular}{ccccc}
    \toprule
     \space & \multicolumn{2}{c}{DS ($\downarrow$)} & \multicolumn{2}{c}{ID Sim ($\uparrow$)} \\
    \midrule
    Dataset     & Learned & Mean & Learned & Mean \\
    \midrule
    FIW         & \textbf{0.6140} & 0.9011 & 0.7003 & \textbf{0.7487} \\
    TSKinFace   & \textbf{0.6780} & 0.9752 & 0.7244 & \textbf{0.7607} \\
    FF-Database & \textbf{0.7078} & 0.9895 & 0.7124 & \textbf{0.7389} \\
    \bottomrule
  \end{tabular}}
  \caption{Performance comparisons of various types of null conditions. The initial row of the table specifies the type of null condition, while the table showcases the performance metrics of diversity (DS) and identity similarity (ID Sim) across different dataset.}
  \label{table:mean-vs-learn}
\end{table}

The experimental results in Table \ref{table:mean-vs-learn} indicate that although mean style latents achieve better identity similarity, they generate less diverse results compared to learned style latents. By using learned style latents as the null condition, a superior trade-off between diversity and identity similarity is achieved across all evaluation datasets. This enhancement is hypothesized to arise from the greater adaptability of learned style latents, which are better equipped to capture variations in attributes such as age and gender, compared to the more constrained mean style latents.

\begin{figure*}
    \centering
    \includegraphics[width=1.0\textwidth]{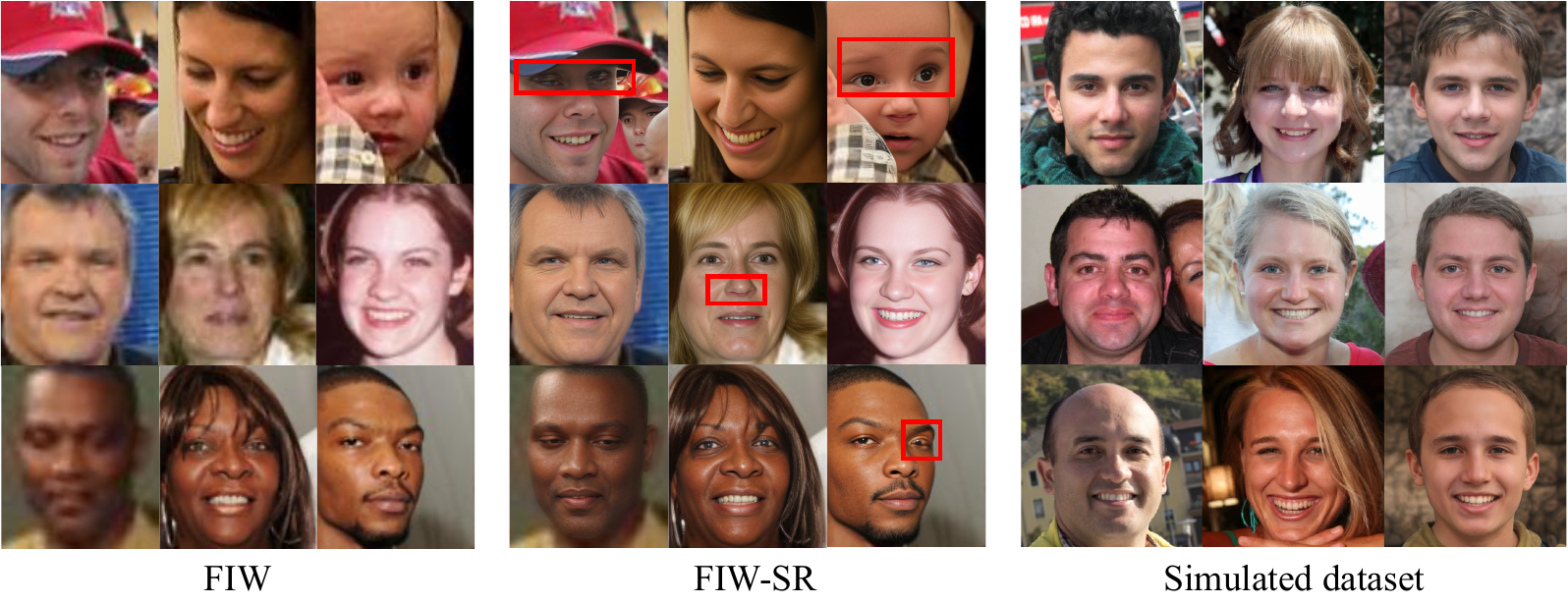}
    \caption{The visual comparison showcases images from the FIW dataset, the FIW dataset after applying super-resolution \cite{zhou2022towards}, and our simulated dataset. In each block, rows represent families, with images of the father, mother, and child displayed from left to right. Red bounding boxes highlight artifacts in the FIW dataset after applying super-resolution.}
    \label{fig:fiw-vs-simulated}
\end{figure*}


\section{More Implementation Details}
\label{sec:sup_implementation}
\subsection{Model Architecture}
Given the success of transformers in processing long vectors in language models, we find them equally well-suited for managing StyleGAN's style latent space. Our model architecture draws inspiration from HyperDiffusion \cite{erkocc2023hyperdiffusion}, which flattens MLP weights into 1D vectors as input of the diffusion model. Our transformer model is designed with an embedding size of 512, 8 layers, and 8 attention heads, effectively handling the complex structure of StyleGAN's latent space.

\subsection{Training Details}
\label{sec:sup_training_details}
We use the AdamW optimizer with a batch size of 1,000 and a learning rate of 0.001, training the model for 4,000 epochs and 500 diffusion timesteps. The training process takes approximately 20 hours on 4 NVIDIA RTX A5000 24GB GPUs.

\subsection{Evaluation Settings}
All evaluation experiments were conducted using a diffusion process with 50 steps. For child prediction, the guidance scale for parents was set to 1.2. For partner prediction, the scale was 1.2 for the child and 0.0 for the parent. Generating a 1024 $\times$ 1024 image takes 0.58 seconds on a single RTX 2080Ti.

\section{Partner Face Synthesis}
\label{sec:sup_partner}
\subsection{Empirical Study}
\label{sec:sup_partner_empirical}
\begin{figure}
    \centering
    \includegraphics[width=1\columnwidth]{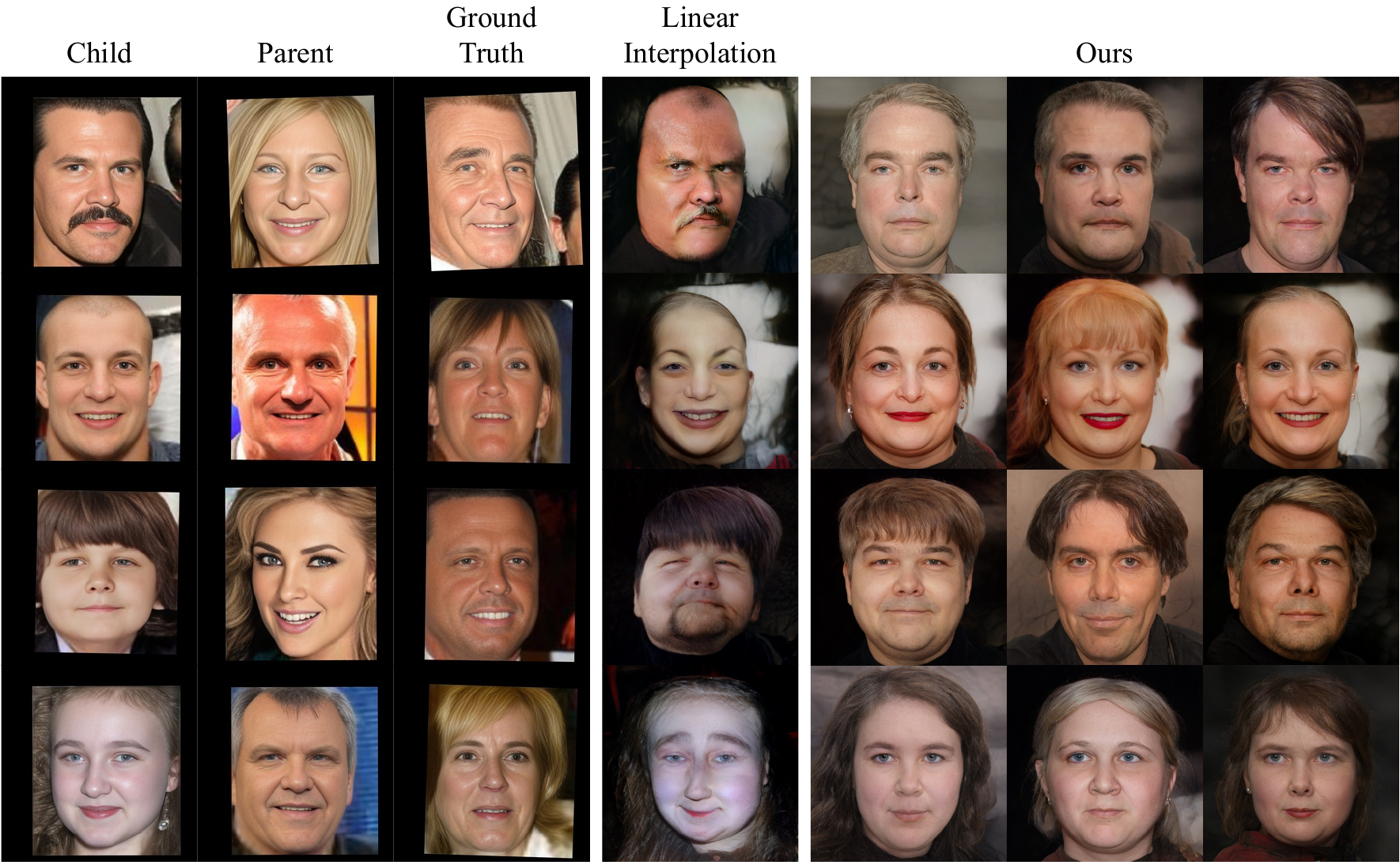}
    \caption{The visual comparison of partner face synthesis.}
    \label{fig:partner-fail-vs-ours}
\end{figure}

Following previous works \cite{kinstyle, stylegene}, a child representation, denoted as $S_{C}$, can be generated through a linear combination of the style latent codes of the father, $S_{F}$, and the mother, $S_{M}$, using an equal weighting (half scale) for both in the style latent space. Formally, this can be expressed as:
\begin{equation}
S_{C} = \frac{1}{2}(S_{F} + S_{M}).
\label{eq:object}
\end{equation}
Additionally, to generate a parent’s representation from the child and the other parent's latents, we could analogically rearrange the equation: 
\begin{equation}
S_{M/F} = (2 \times S_{C} - S_{F/M}).
\label{eq:object}
\end{equation}
However, as illustrated in Figure \ref{fig:partner-fail-vs-ours}, this linear operation often fails to produce optimal results, while our learning-based method achieves superior outcomes in terms of diversity and fidelity. We hypothesize that the failure of this linear approach stems from its reliance on latent extrapolation. Extrapolation involves shifting a latent vector beyond the distribution on which the model was trained, potentially resulting in unpredictable outputs. While StyleGAN can manage controlled extrapolation—such as age progression—reasonably well, extreme shifts beyond the training distribution degrade the quality of the generated results. Notably, while linear operations can achieve child prediction by blending parent latent codes proportionally \cite{kinstyle, stylegene}, partner prediction proves more challenging as it often requires synthesizing representations outside the range of existing latent distributions. For child prediction, interpolation confines the transformation within a range defined by two valid endpoints, ensuring that the output remains realistic and effectively combines characteristics from both parents.

Overall, the inability of linear operations to reliably predict partner representations highlights the necessity of our proposed learning-based method, which addresses these limitations by leveraging data-driven learning to generate more accurate and diverse results.

\begin{table}[t]
    \centering
    \resizebox{\columnwidth}{!}{%
    \setlength{\tabcolsep}{2pt}
    \begin{tabular}{lccccccc}
        \toprule
        \multirow{2}{*}{Methods} & \multicolumn{3}{c}{DS ($\downarrow$)} & \multicolumn{3}{c}{ID Sim ($\uparrow$)} \\
        \cmidrule(lr){2-4} \cmidrule(lr){5-7}
        & FIW & TSKinFace & FF-Database & FIW & TSKinFace & FF-Database \\
        \midrule
        Father       & 0.3243 & 0.3068 & 0.3292 & 0.5678 & 0.5608 & 0.5597 \\
        Mother       & 0.3231 & 0.3166 & 0.3416 & 0.5470 & 0.5787 & 0.5567 \\
        \bottomrule
    \end{tabular}%
    }
    \caption{Quantitative results of diversity (DS) and identity similarity (ID Sim) between predicted partners and their real counterparts.}
    \label{table:partner-similarity}
\end{table}

\subsection{Quantitative Results}
\label{sec:sup_partner_quantitative}
Being pioneers in partner face synthesis, we lack comparable baselines. We use the same evaluation protocol to measure identity similarity between predicted partners and the ground truth partners in Table \ref{table:partner-similarity}. Our results indicate reasonably high accuracy in partner face prediction, both when using a child and father to predict the mother, and when using a child and mother to predict the father.

\section{Fine-grained Age and Gender Control}
\label{sec:sup_attribute_control}
\begin{figure*}
    \centering
    \includegraphics[width=1.0\textwidth]{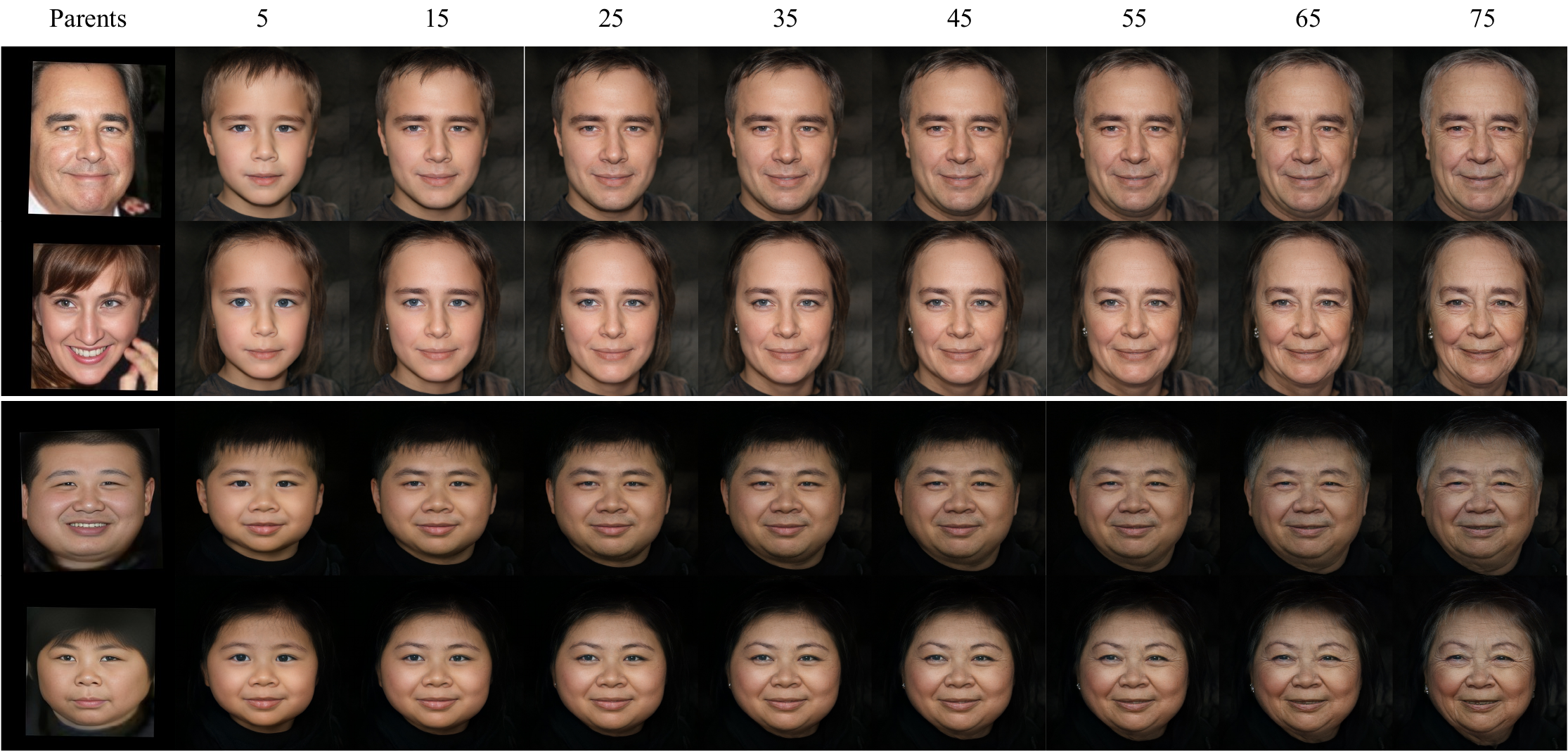}
    \caption{Our framework allows for fine-grained age and gender control, ranging from 0 to 99 years, offering flexibility to generate faces at any arbitrary age within this range.}
    \label{fig:age-and-gender}
\end{figure*}

In Figure \ref{fig:age-and-gender}, the result showcases our proposed method can generate faces with fine-grained age and gender control.

\section{User Study}
We conduct a user study with 100 participants. Our subjective test is authorized by the Academia Sinica IRB committee under the approval number AS-IRB-HS 24031.

\label{sec:sup_user_study}
\begin{figure*}[t!]
    \begin{subfigure}{0.34\textwidth}
        \centering
        \includegraphics[width=\linewidth]{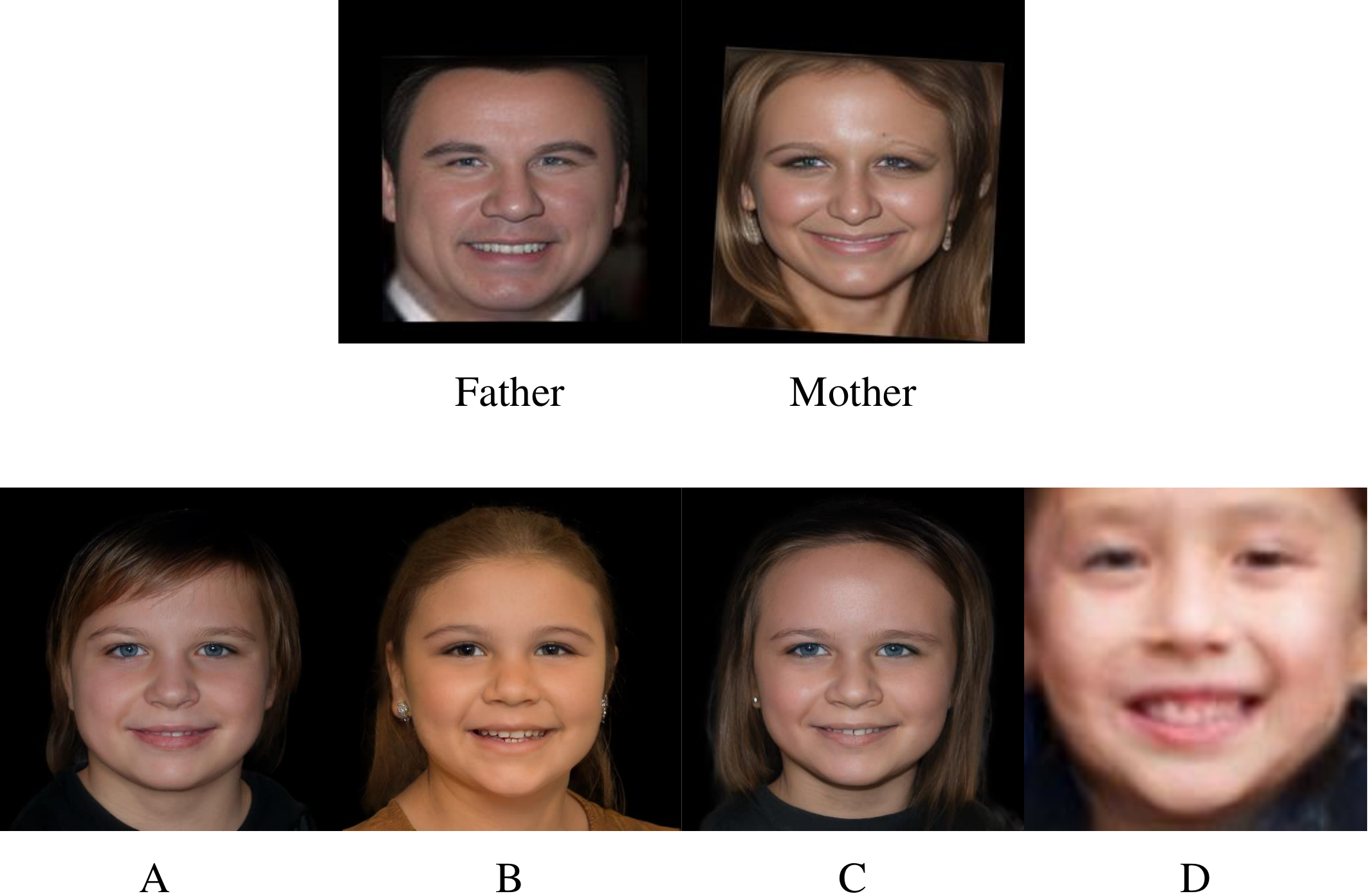}
        \caption{}
        \label{fig:user-subfig1}
    \end{subfigure} 
    \begin{subfigure}{0.34\textwidth}
        \centering
        \includegraphics[width=\linewidth]{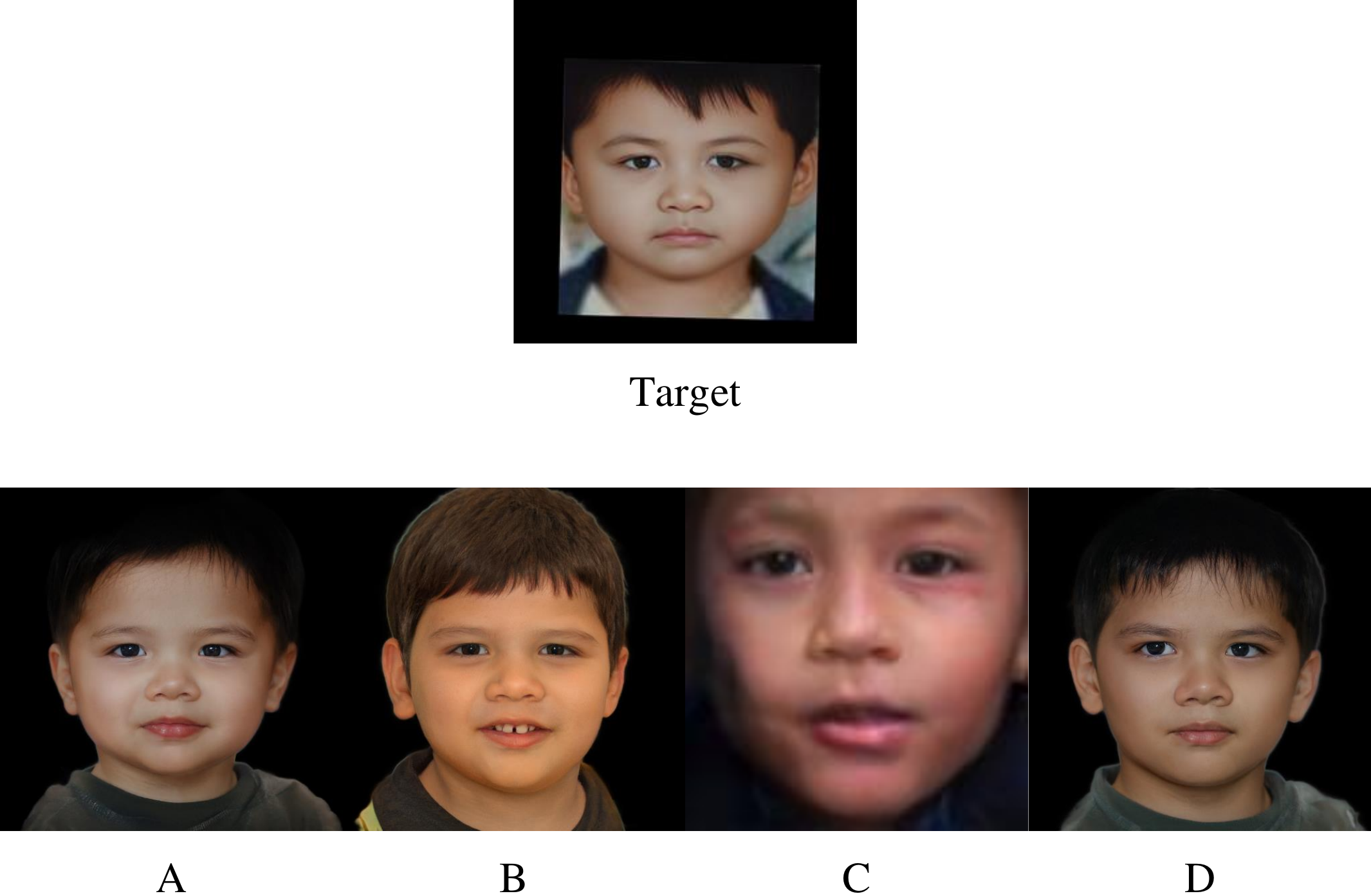}
        \caption{}
        \label{fig:user-subfig2}
    \end{subfigure} 
    \begin{subfigure}{0.34\textwidth}
        \centering
        \includegraphics[width=\linewidth]{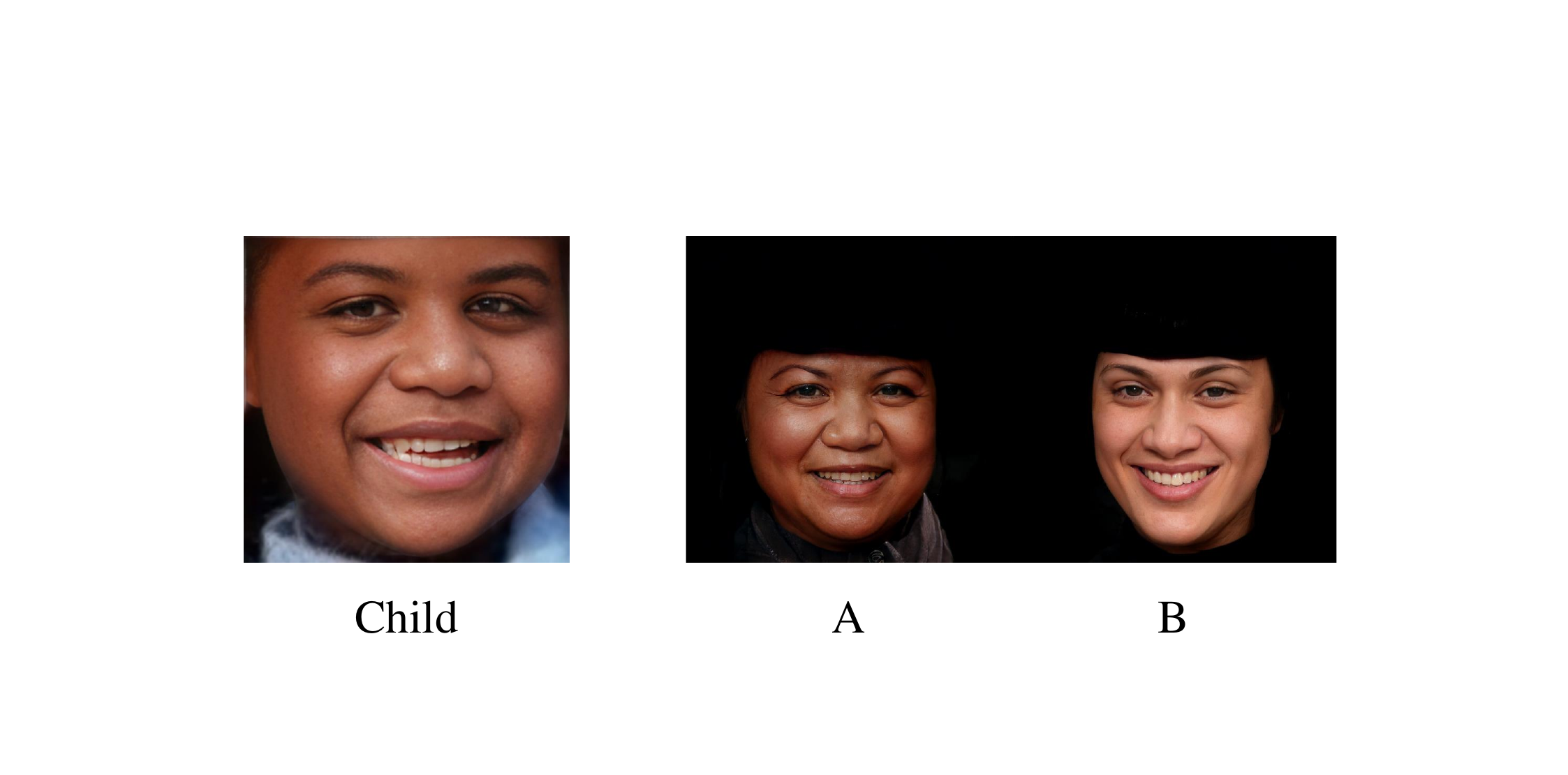}
        \caption{}
        \label{fig:user-subfig3}
    \end{subfigure}
    \caption{Example questions in our user study. (a) An example question in our survey assessing the resemblance of synthesized child faces to the reference face of the parents. (b) An example question in our survey that evaluates the similarity of synthesized child faces to the reference face of the corresponding real child (target). (c) An example question in our survey that asks users to identify the parent (father or mother) based on child face.}
    \label{fig:user-study}
\end{figure*}

\begin{table}[t]
    \centering
    \setlength{\tabcolsep}{4pt} 
    \renewcommand{\arraystretch}{1.2} 
    \begin{adjustbox}{width=0.8\linewidth} 
        \begin{tabular}{lcc}
            \toprule
             & StyleDiT (Ours) & Real \\
            \midrule
            Partner Avg. Score ($\uparrow$) & \textbf{3.84} & 2.84 \\
            \bottomrule
        \end{tabular}
    \end{adjustbox}
    \caption{The average score of partner prediction in the user study.}
    \label{table:partner}
\end{table}

\subsection{Partner Face Synthesis}
In Table \ref{table:partner}, we extend the user study to partner face synthesis, where participants evaluate two parent images (a ground truth and a synthesized image by our model) against a reference photo of a real child. Both images are adjusted to the same age and gender, with evaluators unaware that one image is synthesized.

\subsection{Weighted Average Rank Calculation}
To compute the weighted average rank, we assign the weight of each rank as the number of individuals who selected that rank and then calculate the average weighted by these ranks. For example, if there are four images in a question with ranks ranging from 1 to 4, and an image is ranked as 1 by ten people, 2 by thirty people, 3 by twenty people, and 4 by five people, the weighted average rank for the image is calculated as

\begin{equation}
    \frac{1\times10 + 2\times30 + 3\times20 + 4\times5}{10+30+20+5} = 2.3
\end{equation}

\subsection{Sample Questions and Response Details}
Fig. \ref{fig:user-study} illustrates three example question types from our questionnaire. Fig. \ref{fig:user-subfig1} visually represents a question about the resemblance of a child to parents. Participants are given a pair of parental faces and four options (one from our approach and three from baselines) and are instructed to rank A, B, C, D based on resemblance to the parent images. Similarly, Fig. \ref{fig:user-subfig2} presents another example question in which participants assess the similarity of a child's face and four options (one from our approach and three from baselines), ranked A, B, C, D based on the similarity to the child's image. Additionally, Fig. \ref{fig:user-subfig3} illustrates an example question where participants evaluate the similarity between a child and two options (one generated by our approach and the other representing the real parent), and participants are informed the two options. Participants assign an absolute score (ranging from 1 to 5, with 1 being the least similar and 5 being the most similar) based on the perceived resemblance to the child's face.

\begin{figure}[h]
    \begin{subfigure}{0.5\textwidth}
        \centering
        \includegraphics[width=\linewidth]{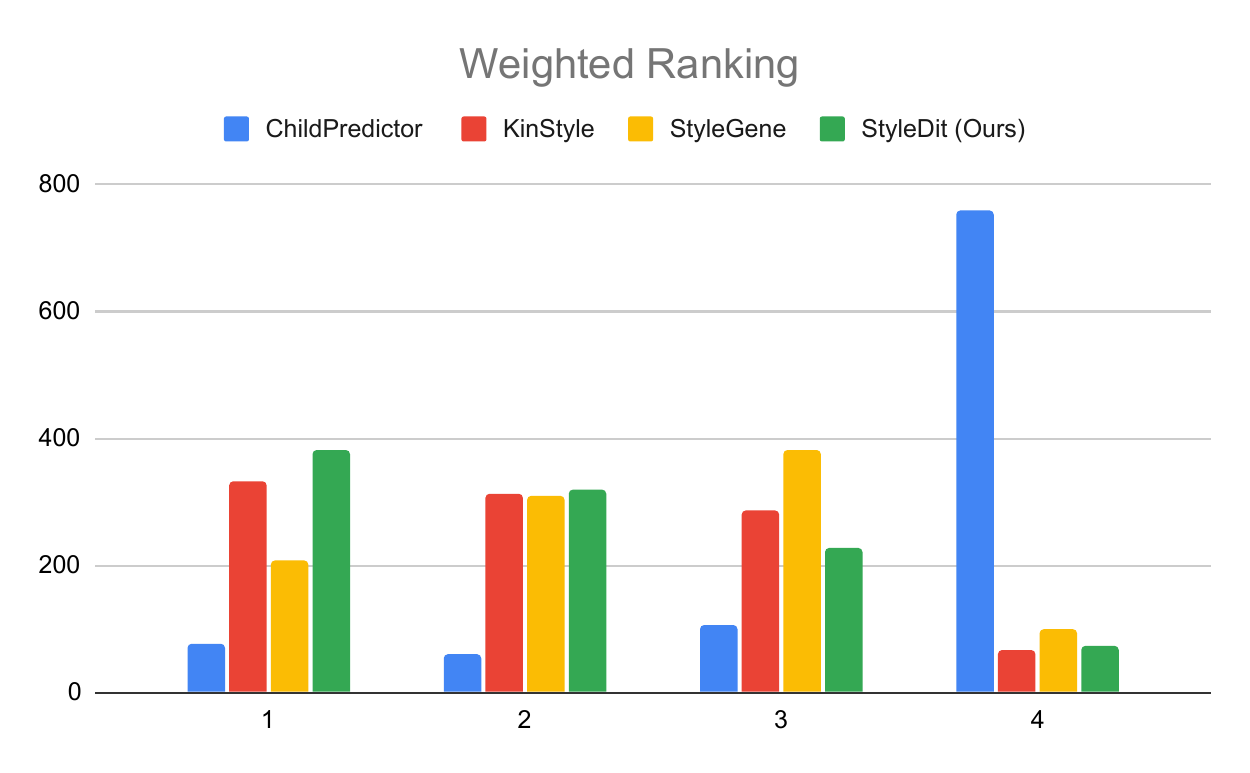}
        \caption{}
        \label{fig:user-rank}
    \end{subfigure} 
    \begin{subfigure}{0.5\textwidth}
        \centering
        \includegraphics[width=\linewidth]{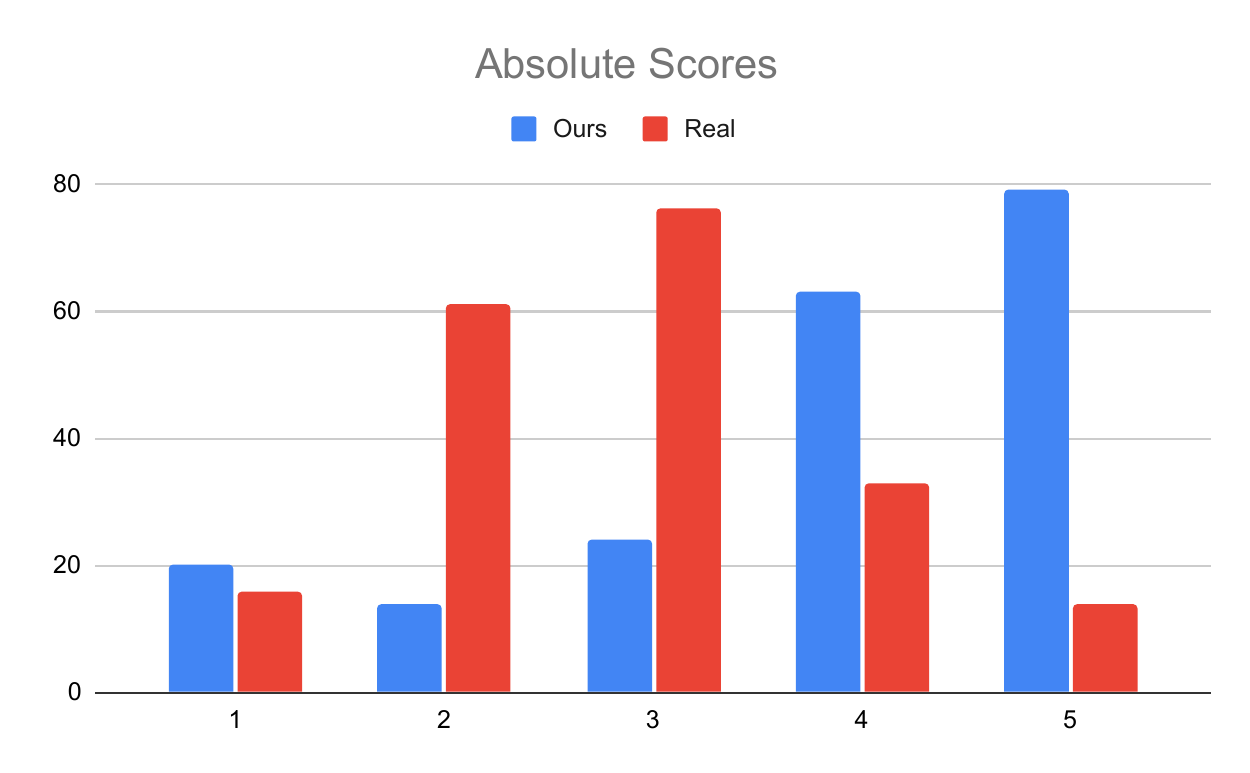}
        \caption{}
        \label{fig:user-pie}
    \end{subfigure}
    \caption{Raw result that we collect from the user study. (a) The weighted ranks in questions regarding parent and child resemblance from the initial and second sessions. (b) The absolute score in questions related to partner similarity from the third session.}
    \label{fig:user-study-chart}
\end{figure}

Figure \ref{fig:user-rank} displays the weighted rank of images in the questions related to resemblance from the first two types of examples, while Figure \ref{fig:user-pie} illustrates the distribution of the absolute scores from the third type of example, related to partner similarity.

\begin{table}[h]
    \centering
    \resizebox{\columnwidth}{!}{%
    \setlength{\tabcolsep}{2pt}
    \begin{tabular}{lccccccc}
        \toprule
        \multirow{2}{*}{Methods} & \multicolumn{3}{c}{DS ($\downarrow$)} & \multicolumn{3}{c}{ID Sim ($\uparrow$)} \\
        \cmidrule(lr){2-4} \cmidrule(lr){5-7}
        & FIW & TSKinFace & FF-Database & FIW & TSKinFace & FF-Database \\
        \midrule
        StyleDiT (Ours) & 0.6140 & 0.6780 & 0.7078 & \textbf{0.7003} & \textbf{0.7244} & \textbf{0.7124} \\
        StyleDiT~$^\dagger$  & 0.5872 & 0.5840 & 0.6119 & 0.6856 & 0.6227 & 0.6092 \\
        StyleDiT~$^\ddagger$ & 0.3578 & 0.3448 & 0.3522 & 0.5840 & 0.5371 & 0.5316 \\
        \bottomrule
    \end{tabular}%
    }
    \caption{Performance comparison of diversity (DS) and identity similarity (ID Sim) between our configuration, fine-tuning, and training solely on the FIW dataset. StyleDiT~$^\dagger$ indicates the fine-tuned version, and StyleDiT~$^\ddagger$ denotes training solely on the real kinship dataset.}
    \label{table:quant-real-data-fiw}
\end{table}

\begin{table}[h]
    \centering
    \resizebox{\columnwidth}{!}{%
    \setlength{\tabcolsep}{2pt}
    \begin{tabular}{lccccccc}
        \toprule
        \multirow{2}{*}{Methods} & \multicolumn{3}{c}{DS ($\downarrow$)} & \multicolumn{3}{c}{ID Sim ($\uparrow$)} \\
        \cmidrule(lr){2-4} \cmidrule(lr){5-7}
        & FIW & TSKinFace & FF-Database & FIW & TSKinFace & FF-Database \\
        \midrule
        StyleDiT (Ours) & 0.6140 & 0.6780 & 0.7078 & \textbf{0.7003} & \textbf{0.7244} & \textbf{0.7124} \\
        StyleDiT~$^\dagger$  & 0.7558 & 0.7839 & 0.7119 & 0.6442 & 0.6861 & 0.6293 \\
        StyleDiT~$^\ddagger$ & 0.3625 & 0.3814 & 0.3361 & 0.5632 & 0.5505 & 0.5239 \\
        \bottomrule
    \end{tabular}%
    }
    \caption{Performance comparison of diversity (DS) and identity similarity (ID Sim) between our configuration, fine-tuning, and training solely on the TSKinFace dataset. StyleDiT~$^\dagger$ indicates the fine-tuned version, and StyleDiT~$^\ddagger$ denotes training solely on the real kinship dataset.}
    \label{table:quant-real-data-tsk}
\end{table}

\begin{table}[h]
    \centering
    \resizebox{\columnwidth}{!}{%
    \setlength{\tabcolsep}{2pt}
    \begin{tabular}{lccccccc}
        \toprule
        \multirow{2}{*}{Methods} & \multicolumn{3}{c}{DS ($\downarrow$)} & \multicolumn{3}{c}{ID Sim ($\uparrow$)} \\
        \cmidrule(lr){2-4} \cmidrule(lr){5-7}
        & FIW & TSKinFace & FF-Database & FIW & TSKinFace & FF-Database \\
        \midrule
        StyleDiT (Ours) & 0.6140 & 0.6780 & 0.7078 & \textbf{0.7003} & \textbf{0.7244} & \textbf{0.7124} \\
        StyleDiT~$^\dagger$  & 0.5625 & 0.6022 & 0.6147 & 0.6817 & 0.6198 & 0.6398 \\
        StyleDiT~$^\ddagger$ & 0.3040 & 0.3133 & 0.3160 & 0.5362 & 0.5181 & 0.5261 \\
        \bottomrule
    \end{tabular}%
    }
    \caption{Performance comparison of diversity (DS) and identity similarity (ID Sim) between our configuration, fine-tuning, and training solely on the FF-Database. StyleDiT~$^\dagger$ indicates the fine-tuned version, and StyleDiT~$^\ddagger$ denotes training solely on the real kinship dataset.}
    \label{table:quant-real-data-ff}
\end{table}

\section{Real Data Utilization}
\label{sec:sup_real_data}
In this section, we present the fine-tuning results and the outcomes of training solely on real data, comparing them to our model, and discuss potential reasons for any shortcomings.

\subsection{Quantitative Results}
For fine-tuning, we utilized each kinship dataset, FIW \cite{fiw}, TSKinFace \cite{TSKinFace}, and FF-Database \cite{childpredictor}, experimenting with 100 epochs and a learning rate of 1e-6. However, as shown in Tables \ref{table:quant-real-data-fiw}, \ref{table:quant-real-data-tsk} and \ref{table:quant-real-data-ff}, the fine-tuned models did not outperform our original setting. While fine-tuning improved diversity, it failed to maintain or even significantly degraded performance in terms of identity similarity. 
Similarly, when training solely on real data with the same kinship datasets and training settings as the original configuration, the results remained consistent: improved diversity came at the cost of a significant decline in identity similarity compared to the original model.
We hypothesize that the observed effects of using a real kinship dataset arise from the inherent complexity of kinship relationships between children and parents in real data, which cannot be fully captured by simple linear interpolation between the style latents of parents. While leveraging real data can produce more diverse results, the limitations imposed by the low quality and scarcity of the dataset significantly affect the generation quality and compromise identity similarity, thereby constraining the overall performance of the model.

\subsection{Discussion}
The results from our quantitative evaluation and user study, as discussed in the main paper, align with previous medical research \cite{debruine_social, alvergne_differential}, further validating the use of our simulated dataset to address the issues of low quality and scarcity in kinship data.

In Figure \ref{fig:fiw-vs-simulated}, we compare the image quality of the FIW dataset, the FIW dataset after applying super-resolution, and our simulated dataset. The FIW dataset faces several limitations, including blurry, low quality, low resolution (108 $\times$ 124), and data scarcity (1,000 families with 1,997 triplets). While super-resolution enhances overall clarity, many facial details, such as wrinkles, textures, and even identity features, remain underdeveloped. Furthermore, it struggles to accurately reconstruct key facial components, such as the eyes. In contrast, our simulated dataset provides significantly better quality, higher resolution (1024 $\times$ 1024), and unlimited data.

\section{Limitation}
Our system relies on GAN inversion, which raises questions about how accurately it can reconstruct images. We can further improve our framework by employing more advanced GAN inversion modules, especially for uses that need precise control over image details and high-quality reconstructions.

\section{Societal Impact}
Our paper introduces methods for creating predictive images of children and partners, serving as a genetic counseling aid for parents to grasp the inheritance of genetic traits. Furthermore, the final output of the method utilizes StyleGAN, whose images are detectable by Deepfake detectors (\url{https://github.com/NVlabs/stylegan3-detector}). Consequently, it does not pose a great risk of social consequences.

\end{document}